\documentclass[12pt, draftclsnofoot, onecolumn]{IEEEtran}
\usepackage[utf8]{inputenc}
\usepackage{bm}
\usepackage{graphicx} 
\usepackage{amsmath}          
\usepackage{cite}
\usepackage{xpatch,amsthm}
\usepackage{listings}
\usepackage[justification=centering,skip=0pt]{caption}
\usepackage{relsize}
\usepackage{subcaption}
\usepackage{verbatim}
\usepackage{amssymb}
\usepackage{mathtools}
\usepackage{dsfont}
\usepackage{hyperref}
\usepackage{xcolor}
\newtheorem{theorem}{Theorem}
\newtheorem{corollary}{Corollary}
\newtheorem{lemma}{Lemma}
\newtheorem{assumption}{Assumption}

\DeclareMathOperator*{\argmin}{arg\,min}
\newcommand{\Lim}[1]{\raisebox{0.5ex}{\scalebox{0.8}{$\displaystyle \lim_{#1}\;$}}}

\ifCLASSINFOpdf
\else
\fi
\begin{document}
\bstctlcite{IEEEexample:BSTcontrol}
%
\title{Over-the-Air Federated Edge Learning with Hierarchical Clustering}
%
%
%

\author{Ozan~Ayg{\"u}n,~\IEEEmembership{Student Member,~IEEE,}
        Mohammad~Kazemi,~\IEEEmembership{Member,~IEEE,}
        Deniz~G{\"u}nd{\"u}z,~\IEEEmembership{Fellow,~IEEE,}
        and~Tolga~M.~Duman,~\IEEEmembership{Fellow,~IEEE}
\thanks{Part of this work was presented at the 2022 IEEE International Conference on Communications (ICC) \cite{aygun2022}.}
\thanks{O. Ayg{\"u}n, M. Kazemi, and T. M. Duman are with the Department of Electrical and Electronics Engineering, Bilkent University, Ankara, 06800, Turkey (email: \{ozan, kazemi, duman\}@ee.bilkent.edu.tr)}
\thanks{Deniz G{\"u}nd{\"u}z is with the Department of Electrical and Electronic Engineering, Imperial College London, London SW7 2BT, U.K.
(e-mail: d.gunduz@imperial.ac.uk)}
\thanks{The authors acknowledge support from TUBITAK through CHIST-ERA project SONATA (CHIST-ERA-20-SICT-004, funded by TUBITAK, Turkey Grant 221N366 and EPSRC, UK Grant EP/W035960/1).

Ozan Ayg{\"u}n's research is also supported by Turkcell A.S. within the framework of 5G and Beyond Joint Graduate Support Programme coordinated by Information and Communication Technologies Authority.}
}

\maketitle

\begin{abstract}
We examine federated learning (FL) with over-the-air (OTA) aggregation, where mobile users (MUs) aim to reach a consensus on a global model with the help of a parameter server (PS) that aggregates the local gradients. In OTA FL, MUs train their models using local data at every training round and transmit their gradients simultaneously using the same frequency band in an uncoded fashion. Based on the received signal of the superposed gradients, the PS performs a global model update. While the OTA FL has a significantly decreased communication cost, it is susceptible to adverse channel effects and noise. Employing multiple antennas at the receiver side can reduce these effects, yet the path-loss is still a limiting factor for users located far away from the PS. To ameliorate this issue, in this paper, we propose a wireless-based hierarchical FL scheme that uses intermediate servers (ISs) to form clusters at the areas where the MUs are more densely located. Our scheme utilizes OTA cluster aggregations for the communication of the MUs with their corresponding IS, and OTA global aggregations from the ISs to the PS. We present a convergence analysis for the proposed algorithm, and show through numerical evaluations of the derived analytical expressions and experimental results that utilizing ISs results in a faster convergence and a better performance than the OTA FL alone while using less transmit power. We also validate the results on the performance using different number of cluster iterations with different datasets and data distributions. We conclude that the best choice of cluster aggregations depends on the data distribution among the MUs and the clusters.
\end{abstract}


%
\IEEEpeerreviewmaketitle

\section{Introduction}
We are surrounded by devices that continuously gather all kinds of information from images, videos, and sound to various sensor measurements. The abundance of generated data has been essential for the rapid advancements in machine learning (ML) in different domains. Traditionally, ML relies on accumulating all the data at a server to train a powerful model with many parameters. However, such centralized training and data accumulation leads to concerns regarding data privacy, communication cost, and latency. Firstly, users are concerned about sharing their personal datasets as it may leak information about the owner beyond the intended use \cite{wei2020federated}. Secondly, offloading collected data samples to a remote sensor, typically for high rate data such as images and videos, requires significant communication resources. Thirdly, applications that need to operate in real-time might be affected by the increased latency since their performance depends on the model response of the simultaneously collected data \cite{lim2020federated}. \textit{Federated learning} (FL) offers an attractive alternative to centralized training, where the training is distributed across user devices, and does not require collecting data at a centralized server \cite{mcmahan2017communication}.

In FL, a parameter server (PS), which keeps track of the global model orchestrates training across a set of mobile users (MUs). At each iteration, the current global model parameters are shared with a subset of the MUs, selected depending on their battery states, computing capabilities, data qualities, or distance to the PS \cite{gunduz2020communicate}. These devices are asked to perform stochastic gradient descent (SGD) on the current model using their local datasets. After completing several local training iterations, each MU sends its model update to the PS. The PS performs model aggregation using these local updates to update the global model, and sends the new model back to a potentially different subset of MUs for the next iteration. Lately, the literature on FL has focused on topics such as data heterogeneity \cite{sery2021over}, privacy \cite{liu2020privacy}, energy efficiency and latency analysis \cite{chen2021joint}. \textit{Federated Averaging} is the simplest and the most popular model aggregation method, which employs simple averaging operation at the PS \cite{konevcny2016federated}. Since the performance mostly depends on the data distribution at the MUs, recent studies focus on heterogeneous datasets across MUs \cite{sery2021over, zhang2021client, zhao2018federated, briggs2020federated, data2021byzantine}. Even though FL aims to protect the local data privacy, some studies show that it is possible to infer user data even from the gradient information, and present approaches to enhance data privacy \cite{liu2020privacy, so2020byzantine, data2021byzantine}. Numerous works have been reported on FL's power consumption and latency, and developed power-efficient FL schemes for MUs with limited power \cite{chen2021joint, dinh2020federated, luo2020hfel}. User heterogeneity is another direction being investigated, where a subset of devices are selected efficiently based on their available power, computing capabilities, and distance \cite{zhang2021client, amiri2021convergence, sun2021dynamic, amiri2021federated, ren2020scheduling}.

While FL among edge devices has a great premise, its implementation in practical wireless scenarios requires combatting adverse channel effects and optimization under limited channel resources. Since the cost of communication in FL depends on the size of the underlying model, the periodic transmission of a large model increases the communication costs and the required bandwidth even though no raw data is being sent. In order to use the bandwidth more efficiently, over-the-air (OTA) aggregation has become a widely used method, where the local updates are sent using the same frequency band, thereby performing computation and transmission simultaneously \cite{amiri2020machine}. However, accurate OTA aggregation requires mitigating adverse effects of wireless channels so that the transmitted signals arrive at the PS at similar power levels. While this is originally achieved by applying channel inversion when accurate channel state information (CSI) is available at the MUs, it is shown in \cite{amiri2021blind} that the effects of wireless fading can also be alleviated by increasing the number of receive antennas at the PS even when the transmitter side has no CSI. Recently, the focus in OTA FL has been on user scheduling \cite{amiri2021convergence, sun2021dynamic, amiri2021federated, ren2020scheduling} and analysis of different wireless channel models \cite{zhu2019broadband, shao2021federated, wei2022federated, amiri2020federated, zhu2020one, amiri2021blind, liu2020privacy}. Other approaches on wireless FL include model compression for the local updates and location-based user scheduling \cite{chen2021, amiri2020machine, chen2021joint, zhu2020one}.

As a possible solution for the increased communication costs, hierarchical federated learning (HFL) has also been proposed. In HFL, intermediate servers (ISs) are employed in areas where the number of MUs is high to form cluster-like structures \cite{abad2020hierarchical}. In this model, the MUs carry out multiple SGD iterations before transmitting their model differences to their corresponding IS. After several cluster aggregation steps between ISs and their corresponding MUs, global aggregation is carried out at the PS, using the IS cluster updates. Studies on HFL focus on the system performance when the users have non-independent and identically distributed (non-i.i.d.) data distributions \cite{briggs2020federated}, power, latency, and convergence analysis \cite{liu2020client, abad2020hierarchical, liu2021hierarchical, wang2020local}, and optimal resource allocation schemes \cite{luo2020hfel}. 

In this paper, we study an HFL scenario, similarly to the one in \cite{abad2020hierarchical}, but we consider OTA aggregation both at the ISs and the PS.
Our specific contributions in this framework are as follows:

\begin{itemize}
    \item We provide the system model and channel specifications for the considered wireless hierarchical federated learning (W-HFL) system including the intra-cluster and inter-cluster interference effects. We present the details of the proposed OTA aggregation algorithm.
    \item We conduct a detailed convergence analysis for the proposed model, where the effects of interference and noise terms can be clearly identified. We also provide an upper bound on the convergence rate and numerically compare the convergence rate with that of the conventional FL, where all the MUs communicate with PS directly without the need for an IS. We show through numerical evaluations of the analytical results that the proposed algorithm has a higher convergence rate than conventional FL, and has a competitive performance compared to the baseline scheme with error-free links.
    \item We demonstrate via experimental results on MNIST and CIFAR-10 datasets with different data distributions that the proposed scheme exhibit faster convergence behavior and converges to a a more reliable model compared to that of the conventional FL while also using less power at the edge. 
\end{itemize}

The rest of the paper is organized as follows. In Section \ref{sec:systemmodel}, we introduce the learning objective as well as the structure of W-HFL. In Section \ref{sec:whfl}, we provide the communication model of the proposed algorithm. In Section \ref{sec:convanalysis}, the convergence analysis of W-HFL is presented, and it is upper-bounded under some convexity assumptions. In Section \ref{sec:simresults}, we give experimental and numerical results to compare our algorithm with the conventional FL as well as the baseline approaches, and we conclude the paper in Section \ref{sec:conclusion}.

\section{System Model} \label{sec:systemmodel}
\begin{figure}
\begin{center}
\includegraphics[width=0.4\columnwidth]{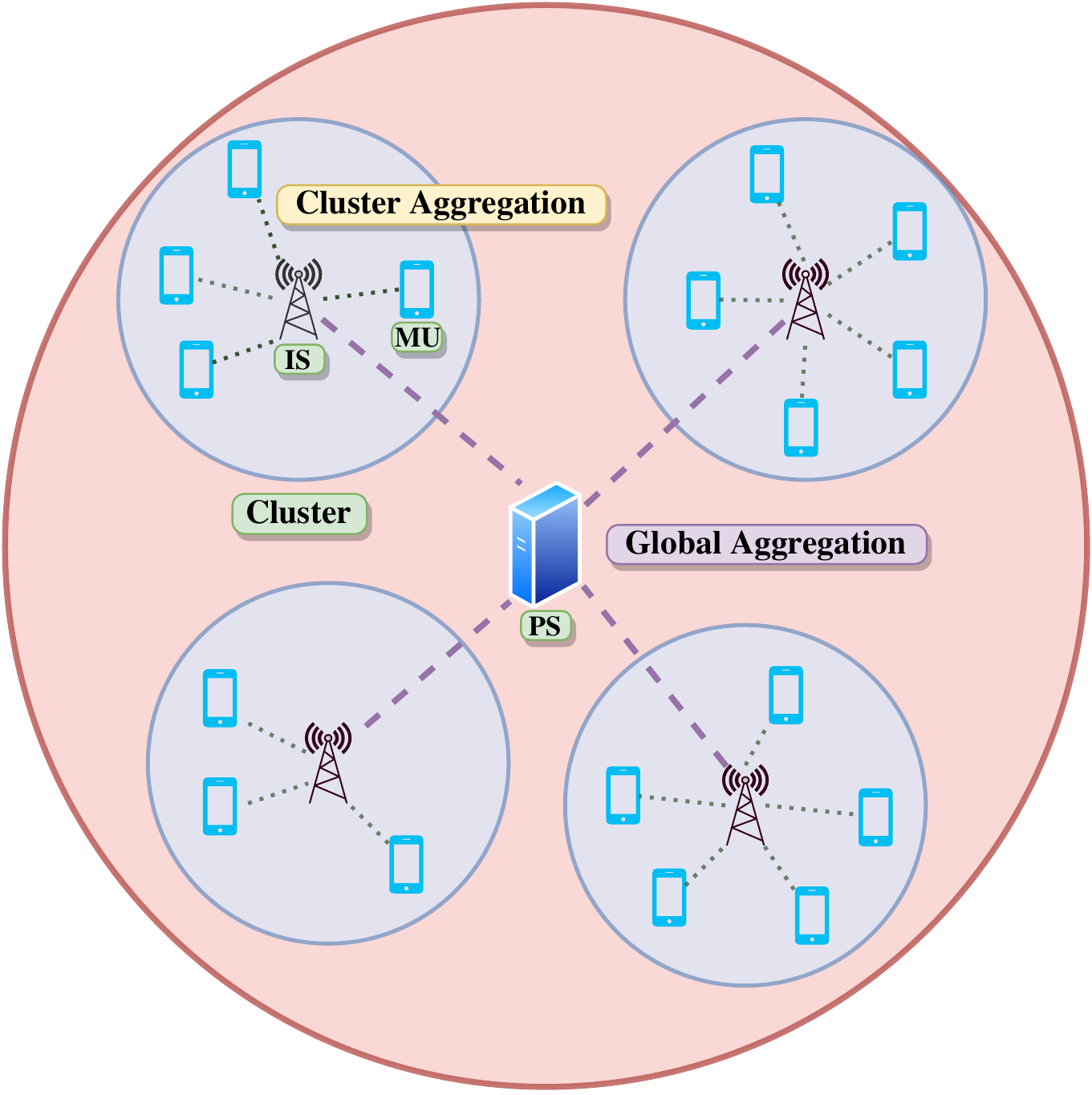} 
\caption{Illustration of the W-HFL system model}
\label{fig:system}
\end{center}
\end{figure}

The objective of W-HFL is to minimize a loss function $F(\bm{\theta})$ with respect to the model weight vector $\bm{\theta} \in \mathbb{R}^{2N}$, where $2N$ is the model dimension. Our system consists of $C$ clusters each containing an IS and $M$ MUs, and a PS as depicted in Fig. \ref{fig:system}.
The dataset of the $m$-th MU in the $c$-th cluster is denoted as $\mathcal{B}_{c,m}$, and we define $B \triangleq \sum_{c = 1}^{C} \sum_{m = 1}^{M} |\mathcal{B}_{c,m}|$. We have
\begin{equation}
    F(\bm{\theta}) = \sum_{c = 1}^{C} \sum_{m = 1}^{M} \frac{|\mathcal{B}_{c,m}|}{B} F_{c,m}(\bm{\theta}),
    \label{eq:empiricalloss}
\end{equation}
where 
    $F_{c,m}(\bm{\theta}) \triangleq \frac{1}{|\mathcal{B}_{c,m}|} \sum_{u \in \mathcal{B}_{c,m}} f(\bm{\theta},u)$,
with $f(\bm{\theta},u)$ denoting the loss function corresponding to parameter vector $\bm{\theta}$ and data sample $u$.

We consider a hierarchical and iterative approach consisting of global, cluster, and user iterations to minimize \eqref{eq:empiricalloss}. In every cluster iteration, the MUs carry out $\tau$ user iterations using their local datasets, then send their model updates to their corresponding ISs for cluster aggregation. $I$ cluster iterations are performed at each IS before all the updated models are forwarded to the PS for global aggregation. Consider the $j$-th user iteration of the $i$-th cluster iteration of the $t$-th global iteration by the $m$-th user in the $c$-th cluster. The weight update is performed employing SGD as follows:
\begin{equation}
    \bm{\theta}_{c,m}^{i,j+1}(t) = \bm{\theta}_{c,m}^{i,j}(t) - \eta_{c,m}^{i,j}(t) \nabla F_{c,m} (\bm{\theta}_{c,m}^{i,j}(t), \bm{\xi}_{c,m}^{i,j}(t)),
\end{equation}
where $\eta_{c,m}^{i,j}(t)$ is the learning rate, $\nabla F_{c,m} (\bm{\theta}_{c,m}^{i,j}(t), \bm{\xi}_{c,m}^{i,j}(t))$ denotes the stochastic gradient estimate for the weight vector $\bm{\theta}_{c,m}^{i,j}(t)$ and a randomly sampled batch of data samples $\bm{\xi}_{c,m}^{i,j}(t)$ sampled from the dataset $\mathcal{B}_{c,m}$. Initially, $\bm{\theta}_{c,m}^{1,1}(t) = \bm{\theta}_{IS,c}^{i}(t), \forall i \in [I]$, where $[I] \triangleq \{ 1,2,\ldots,I \}$, and $\bm{\theta}_{IS,c}^{1}(t) = \bm{\theta}_{PS}(t)$, where $\bm{\theta}_{PS}(t)$ is the global model at the PS at the $t$-th global iteration and $\bm{\theta}_{IS,c}^{i}(t)$ denotes the local model of the IS in the $c$-th cluster at the $i$-th cluster iteration. The purpose of employing ISs is to accumulate the local model differences within each cluster more frequently over smaller areas before obtaining the global model $\bm{\theta}_{PS}(t)$ for the next global iteration. Also, note that $\nabla F_{c,m} (\bm{\theta}_{c,m}^{i,j}(t), \bm{\xi}_{c,m}^{i,j}(t))$ is an unbiased estimator of $\nabla F_{c,m} (\bm{\theta}_{c,m}^{i,j}(t))$, i.e., $\mathbb{E}_{\xi} \left[ \nabla F_{c,m} (\bm{\theta}_{c,m}^{i,j}(t), \bm{\xi}_{c,m}^{i,j}(t)) \right] = \nabla F_{c,m} (\bm{\theta}_{c,m}^{i,j}(t)),$
where the expectation is over the randomness due to SGD.


\section{Wireless Hierarchical Federated Learning (W-HFL)} \label{sec:whfl}
\subsection{Ideal Communication} 
We first consider the case in which all the communications among all the units is error-free as a benchmark. In this case, after $\tau$ user iterations, each MU calculates its model difference to be sent to its corresponding IS as
\begin{equation}
    \Delta \bm{\theta}_{c,m}^{i}(t) = \bm{\theta}_{c,m}^{i,\tau+1}(t) - \bm{\theta}_{IS,c}^{i}(t).
\end{equation}
Then, the cluster aggregation at the $c$-th cluster is performed as
\begin{equation}
    \bm{\theta}_{IS,c}^{i+1}(t) = \bm{\theta}_{IS,c}^{i}(t) + \frac{1}{M} \sum_{m = 1}^{M} \Delta \bm{\theta}_{c,m}^{i}(t).
\end{equation}
After completing $I$ cluster iterations in each cluster, ISs send their model differences to the PS, which can be written as
\begin{equation}
    \Delta \bm{\theta}_{PS,c}(t) = \bm{\theta}_{IS,c}^{I+1}(t) - \bm{\theta}_{PS}(t).
    \label{global1}
\end{equation}
The global update rule is
$\Delta \bm{{\theta}}_{PS}(t) = \frac{1}{C} \sum_{c = 1}^{C} \Delta \bm{\theta}_{PS,c}(t)$.
Using recursion, we can conclude that
\begin{equation}
    \Delta \bm{\theta}_{PS}(t) = \frac{1}{MC} \sum_{c = 1}^{C} \sum_{m = 1}^{M} \sum_{i = 1}^{I} \Delta \bm{\theta}_{c,m}^{i}(t).
    \label{eq: globalmodeldiff}
\end{equation}
After the global aggregation, the model at the PS is updated as
$\bm{\theta}_{PS}(t+1) = \bm{\theta}_{PS}(t) + \Delta \bm{\theta}_{PS}(t)$.
\subsection{OTA Communication}
We now introduce the scheme referred to as OTA communications to be used for all the links from the users to the ISs, and from the ISs to the PS. Since model differences are transmitted via a common wireless medium in both cluster and global updates, estimated versions of $\Delta \bm{\theta}_{IS,c}(t)$ and $\Delta \bm{\theta}_{PS}(t)$ are received at the ISs and the PS, where the system noise and inter/intra cluster interference are present. In our setup, ISs and PS have $K$ and $K^{\prime}$ receive antennas, respectively, while both ISs and the MUs are equipped with a single transmit antenna \footnote{For the case of multiple transmit antennas at the ISs, as long as each IS transmits the weighted and phase shifted versions of the same stream, i.e., employs beamforming, the same setup is applicable.}. Also, we assume perfect channel state information (CSI) at the receiver ends.
\subsubsection{Cluster Aggregation} 
In OTA communication, the local updates $\Delta \bm{\theta}_{c,m}^{i}(t) \in \mathbb{R}^{2N}$ are sent without any coding. In order to increase the spectral efficiency, the model differences are grouped to form a complex vector $\Delta \bm{\theta}_{c,m}^{i,cx}(t) \in \mathbb{C}^{N}$ with entries $\Delta \theta_{c,m}^{i,n,cx}(t)$ for $m \in [M], c \in [C], i \in [I]$, with the following real and imaginary parts
\begin{subequations}\label{eq:reco}
    \begin{alignat}{2}
        & \Delta \bm{\theta}_{c,m}^{i,re}(t) \triangleq \big[ \Delta \theta_{c,m}^{i,1}(t), \theta_{c,m}^{i,2}(t), \ldots, \Delta \theta_{c,m}^{i,N}(t) \big]^{T}, \label{sub-eq-1:reco} \\
        & \Delta \bm{\theta}_{c,m}^{i,im}(t) \triangleq  \big[ \Delta \theta_{c,m}^{i,N+1}(t), \theta_{c,m}^{i,N+2}(t), \ldots, \Delta \theta_{c,m}^{i,2N}(t) \big]^{T}, \label{sub-eq-2:reco} 
    \end{alignat}
\end{subequations}
where $\Delta \theta_{c,m}^{i,n}(t)$ denotes the $n$-th entry of $\Delta \bm{\theta}_{c,m}^{i}(t)$ for $n \in [2N]$. The resulting complex vector is transmitted through the wireless medium. The received signal at the $k$-th antenna of the $c$-th IS in the $i$-th cluster iteration can be represented as
\begin{equation} \label{eq: ISreceivedsignal}
    \bm{y}_{IS,c,k}^{i}(t) = P_{t} \sum_{c^{\prime} = 1}^{C} \sum_{m = 1}^{M} \bm{h}_{c^{\prime},m,c,k}^{i}(t) \circ \Delta \bm{\theta}_{c^{\prime},m}^{i,cx}(t) + \bm{z}_{IS,c,k}^{i}(t),
\end{equation}
where $P_{t}$ is the power multiplier at the $t$-th global iteration, $\circ$ denotes the element-wise (Hadamard) product, $\bm{z}_{IS,c,k}^{i}(t) \! \in \! \mathbb{C}^{N}$ is the circularly symmetric additive white Gaussian noise (AWGN) vector with i.i.d. entries with zero mean and variance of $\sigma_{z}^{2}$; i.e., $ z_{IS,c,k}^{i,n}(t) \! \sim \! \mathcal{CN}(0,\sigma_{z}^{2}), n \! \in \! \big[ N \big]$. $\bm{h}_{c^{\prime},m,c,k}^{i}(t) \! \in \! [N]$ is the channel coefficient vector between the $m$-th MU in the $c^{\prime}$-th cluster and the $c$-th IS, whose $n$-th entry is modelled as $h_{c^{\prime},m,c,k}^{i,n}(t) \! = \! \sqrt{\beta_{c^{\prime},m,c}} g_{c^{\prime},m,c,k}^{i,n}(t)$, where $g_{c^{\prime},m,c,k}^{i,n}(t) \! \sim \! \mathcal{CN}(0,\sigma_{h}^{2})$ is the small-scale fading coefficient (i.e., Rayleigh fading), and $\beta_{c^{\prime},m,c}$ is the large-scale fading coefficient modeled as $\beta_{c^{\prime},m,c} \! = \! \left( d_{c^{\prime},m,c} \right)^{-p}$, where $p$ represents the path-loss exponent and $d_{c^{\prime},m,c}$ is the distance between the $m$-th user in the $c^{\prime}$-th cluster and the $c$-th IS. 

Knowing the CSI perfectly, the $c$-th IS combines the received signals as
\begin{equation} \label{eq: IScombining}
    \bm{y}_{IS,c}^{i}(t) = \frac{1}{K} \sum_{k = 1}^{K}  \Big( \sum_{m = 1}^{M} \bm{h}_{c,m,c,k}^{i}(t) \Big)^{\ast} \circ \bm{y}_{IS,c,k}^{i}(t),
\end{equation}
whose $n$-th entry can be written as
\begin{equation} \label{eq: IScombiningsymbol}
    y_{IS,c}^{i,n}(t) = \frac{1}{K} \sum_{k = 1}^{K}  \Big( \sum_{m = 1}^{M} h_{c,m,c,k}^{i,n}(t) \Big)^{\ast}  y_{IS,c,k}^{i,n}(t),
\end{equation}
where $y_{IS,c,k}^{i,n}(t)$ denotes the $n$-th entry of $\bm{y}_{IS,c,k}^{i}(t), n \in [N]$. Substituting (\ref{eq: ISreceivedsignal}) into (\ref{eq: IScombining}), and using (\ref{eq: IScombiningsymbol}), we get
\begin{alignat}{2}
        y_{IS,c}^{i,n}(t)  & =  \underbrace{\frac{P_{t}}{K} \sum_{m = 1}^{M}  \Big( \sum_{k = 1}^{K} \lvert h_{c,m,c,k}^{i,n}(t) \rvert^{2} \Big)  \Delta \theta_{c,m}^{i,n,cx}(t)}_{\text{$y_{IS,c}^{i,n,sig}(t)$ (signal term)}} + \underbrace{  \frac{1}{K} \sum_{m = 1}^{M} \sum_{k = 1}^{K}  (h_{c,m,c,k}^{i,n}(t))^{\ast} z_{IS,c,k}^{i,n}(t)}_{\text{$y_{IS,c}^{i,n,no}(t)$ (noise term)}} \nonumber \\
        & \hspace{0.3cm} + \frac{P_{t}}{K} \sum_{m = 1}^{M} \sum_{k = 1}^{K} (h_{c,m,c,k}^{i,n}(t))^{\ast} \! \bigg( \!\!\! \underbrace{\sum_{\substack{m^{\prime} = 1 \\ m^{\prime} \neq m}}^{M} \!\! h_{c,m^{\prime},c,k}^{i,n}(t) \Delta \theta_{c,m^{\prime}}^{i,n,cx}(t)}_{\text{$ y_{IS,c}^{i,n,int1}(t)$ (Intra-cluster interference)}} \!\! + \! \underbrace{\sum_{\substack{c^{\prime} = 1 \\ c^{\prime} \neq c}}^{C} \sum_{m^{\prime} = 1}^{M}  h_{c^{\prime},m^{\prime},c,k}^{i,n}(t)  \Delta \theta_{c^{\prime},m^{\prime}}^{i,n,cx}(t)}_{\text{$ y_{IS,c}^{i,n,int2}(t)$ (Inter-cluster interference)}}  \bigg) \nonumber \\
        & = y_{IS,c}^{i,n,sig}(t) + y_{IS,c}^{i,n,int1}(t) + y_{IS,c}^{i,n,int2}(t) + y_{IS,c}^{i,n,noise}(t). 
        \label{eq: ISreceivedlong}
\end{alignat}
Aggregated model differences for $n \in [N]$, can be recovered by
\begin{equation} \label{eq:sin11}
    \Delta \hat{\theta}_{IS,c}^{i,n}(t) = \frac{1}{P_{t} M \sigma_{h}^{2} \Bar{\beta}_{c}} \operatorname{Re}\{ y_{IS,c}^{i,n}(t) \}, \hspace{1cm} \Delta \hat{\theta}_{IS,c}^{i,n+N}(t) = \frac{1}{P_{t} M \sigma_{h}^{2} \Bar{\beta}_{c}} \operatorname{Im}\{ y_{IS,c}^{i,n}(t) \},
\end{equation}
where $\Bar{\beta}_{c} = \sum_{m = 1}^{M} \beta_{c,m,c}$, and $\operatorname{Re} \{ a \}$ and $\operatorname{Im} \{ a \}$ denote the real and imaginary parts of $a$, respectively. Finally, the cluster model update can be written as
\begin{equation}
    \bm{\theta}_{IS,c}^{i+1}(t) = \bm{\theta}_{IS,c}^{i}(t) + \Delta \bm{\hat{\theta}}_{IS,c}^{i}(t),
\end{equation}
where $\Delta \bm{\hat{\theta}}_{IS,c}^{i}(t) \triangleq \big[ \Delta \hat{\theta}_{IS,c}^{i,1}(t)  \Delta\hat{\theta}_{IS,c}^{i,2}(t)  \ldots  \Delta\hat{\theta}_{IS,c}^{i,2N}(t) \big]^{T}$. 
\subsubsection{Global Aggregation}
Global aggregation is very similar to cluster aggregation, where each IS has a single transmit antenna and the PS has $K^{\prime}$ receive antennas. After $I$ cluster iterations are completed to obtain the signal to be transmitted from the $c$-th IS, model differences are grouped to form a complex vector $\Delta \bm{\theta}_{PS,c}^{cx} \in \mathbb{C}^{N}$, with the following real and imaginary parts
\begin{subequations}\label{eq:reim}
    \begin{alignat}{2}
        & \!\!\Delta \bm{\theta}_{PS,c}^{re}(t) \triangleq \big[ \Delta \theta_{PS,c}^{1}(t), \theta_{PS,c}^{2}(t), \ldots, \Delta \theta_{PS,c}^{N}(t) \big]^{T}, \label{sub-eq-1:reim} \\
        & \!\!\Delta \bm{\theta}_{PS,c}^{im}(t) \!\triangleq \big[ \Delta \theta_{PS,c}^{N+1}(t), \theta_{PS,c}^{N+2}(t), \ldots, \Delta \theta_{PS,c}^{2N}(t) \big]^{T}, \label{sub-eq-2:reim} 
    \end{alignat}
\end{subequations}
where $\Delta \theta_{PS,c}^{n}(t)$ denotes the $n$-th gradient value at the $c$-th IS. The received signal at the $k^{\prime}$-th antenna of the PS can be written as
\begin{equation}
    \bm{y}_{PS,k^{\prime}}(t) = P_{IS,t} \sum_{c = 1}^{C} \bm{h}_{PS,c,k^{\prime}}(t) \circ \Delta \bm{\theta}_{PS,c}^{cx}(t) + \bm{z}_{PS,k^{\prime}}(t),
\end{equation}
where $P_{IS,t}$ is the power multiplier of the $c$-th IS at the $t$-th global iteration, $\bm{z}_{PS,k^{\prime}}(t) \in \mathbb{C}^{N}$ is the circularly symmetric AWGN noise with i.i.d. entries with zero mean and variance $\sigma_{z}^{2}$; i.e., $ z_{PS,k^{\prime}}^{n}(t) \sim \mathcal{CN}(0,\sigma_{z}^{2})$. The channel coefficient between the $c$-th IS and the PS is modelled as $\bm{h}_{PS,c,k^{\prime}}(t) = \sqrt{\beta_{IS,c}} ~ \bm{g}_{PS,c,k^{\prime}}(t)$, where $\bm{g}_{PS,c,k^{\prime}}(t) \in \mathbb{C}^{N}$ is the small-scale fading coefficient vector with entries $g_{PS,c,k^{\prime}}^{n}(t) \sim \mathcal{CN}(0,\sigma_{h}^{2})$, $\beta_{IS,c}$ is the large-scale fading coefficient modeled as $\beta_{IS,c} = \big( d_{IS,c} \big)^{-p}$, where $d_{IS,c}$ denotes the distance between the $c$-th IS and the PS. 

Knowing the CSI perfectly, the received signal at the PS is combined as
\begin{equation}
    \bm{y}_{PS}(t) \triangleq \frac{1}{K^{\prime}} \sum_{k^{\prime} = 1}^{K^{\prime}} \Big( \sum_{c = 1}^{C} \bm{h}_{PS,c,k^{\prime}}(t) \Big)^{\ast} \circ \bm{y}_{PS,k^{\prime}}(t).
\end{equation}
Estimated global model differences at the PS can be recovered as
\begin{equation} \label{eq:sin22}
    \Delta \hat{\theta}_{PS}^{n}(t) = \frac{1}{P_{IS,t} C \sigma_{h}^{2} \Bar{\beta}} \operatorname{Re}\{ y_{PS}^{n}(t) \}, \hspace{1cm} \Delta \hat{\theta}_{PS}^{n+N}(t) = \frac{1}{P_{IS,t} C \sigma_{h}^{2} \Bar{\beta}} \operatorname{Im}\{ y_{PS}^{n}(t) \},
\end{equation}
where $\Bar{\beta} = \sum_{c = 1}^{C} \beta_{IS,c}$. Finally, the global aggregation is performed using
\begin{align} 
    \bm{\theta}_{PS}(t+1) = \bm{\theta}_{PS}(t) + \Delta \bm{\hat{\theta}}_{PS}(t), \label{eq:globalaggregation}
\end{align}
where $\Delta \bm{\hat{\theta}}_{PS}(t) = \big[ \Delta \hat{\theta}_{PS}^{1}(t)  \Delta\hat{\theta}_{PS}^{2}(t)  \ldots  \Delta\hat{\theta}_{PS}^{2N}(t) \big]^{T}$.

The $n$-th symbol can be written as
\begin{subequations}\label{eq:sin2}
    \begin{alignat}{2}
        y_{PS}^{n}(t) & = \frac{1}{K^{\prime}} \sum_{k^{\prime} = 1}^{K^{\prime}}  \Big( \sum_{c = 1}^{C} h_{PS,c,k^{\prime}}(t) \Big)^{\ast} y_{PS,k^{\prime}}^{n}(t) \\
        & = \underbrace{ P_{IS,t} \sum_{c = 1}^{C} \Big( \frac{1}{K^{\prime}} \sum_{k^{\prime} = 1}^{K^{\prime}} | h_{PS,c,k^{\prime}}^{n}(t) |^{2} \Big) \Delta \theta_{PS,c}^{n,cx}(t)}_{\text{Signal Term}} \\
        & \hspace{0.3cm} + \underbrace{ \frac{P_{IS,t}}{K^{\prime}} \! \sum_{c = 1}^{C} \sum_{\substack{c^{\prime} = 1 \\ c^{\prime} \neq c}}^{C} \sum_{k^{\prime} = 1}^{K^{\prime}} \! \big( h_{PS,c,k^{\prime}}^{n}(t) \big)^{\ast} h_{PS,c^{\prime},k^{\prime}}^{n}(t) \Delta \theta_{PS,c^{\prime}}^{n,cx}(t)}_{\text{Interference Term}} \! + \! \underbrace{ \frac{1}{K^{\prime}} \! \sum_{c = 1}^{C} \sum_{k^{\prime} = 1}^{K^{\prime}} \! \big( h_{PS,c,k^{\prime}}^{n}(t) \big)^{\ast} z_{PS,k^{\prime}}^{n}(t)}_{\text{Noise Term}} \nonumber \\
        & = y_{PS}^{n,sig}(t) + y_{PS}^{n,int}(t) + y_{PS}^{n,noise}(t). \label{sub-eq-3:sin2} 
    \end{alignat}
\end{subequations}
Since we can write
    $\Delta \theta_{PS,c}^{n,cx}(t) = \Delta \theta_{PS,c}^{n}(t) + j \Delta \theta_{PS,c}^{n+N}(t)$,
using (\ref{global1}) and recursively adding previous cluster iterations, we obtain
\begin{align}
    \Delta \theta_{PS,c}^{n,cx}(t) & = \big( \Delta \theta_{IS,c}^{I+1,n}(t) - \Delta \theta_{IS,c}^{1,n}(t) \big)  + j \big( \Delta \theta_{IS,c}^{I+1,n+N}(t) - \Delta \theta_{IS,c}^{1,n+N}(t) \big) \\
    &= \sum_{i = 1}^{I} \Delta \hat{\theta}_{IS,c}^{i,n}(t) + j \Delta \hat{\theta}_{IS,c}^{i,n+N}(t) \\
    &= \frac{1}{P_{t} M \sigma_{h}^{2} \Bar{\beta}_{c}} \sum_{i = 1}^{I} y_{IS,c}^{i,n}(t).
    \label{is_modeldiff} 
\end{align}
Substituting Equation \eqref{is_modeldiff} into \eqref{eq:sin2}, we have
\begin{align} \label{eq:ps_received}
    & y_{PS}^{n}(t) \!\! = \!\! P_{IS,t} \! \sum_{c = 1}^{C} \! \bigg( \! \frac{1}{K^{\prime}} \! \sum_{k^{\prime} = 1}^{K^{\prime}} \! | h_{PS,c,k^{\prime}}^{n}(t) |^{2} \! \bigg) \!\! \bigg( \! \frac{1}{P_{t} M \sigma_{h}^{2} \Bar{\beta}_{c}} \! \sum_{i = 1}^{I} y_{IS,c}^{i,n}(t) \!\! \bigg) \! + \! \frac{1}{K^{\prime}} \! \sum_{c = 1}^{C} \sum_{k^{\prime} = 1}^{K^{\prime}} \big( \! h_{PS,c,k^{\prime}}^{n}(t) \! \big)^{\ast} z_{PS,k^{\prime}}^{n}(t) \nonumber \\
    & \hspace{1.5cm} + \frac{P_{IS,t}}{K^{\prime}} \sum_{c = 1}^{C} \sum_{\substack{c^{\prime} = 1 \\ c^{\prime} \neq c}}^{C^{\prime}} \sum_{k^{\prime} = 1}^{K^{\prime}} \big( h_{PS,c,k^{\prime}}^{n}(t) \big)^{\ast} h_{PS,c^{\prime},k^{\prime}}^{n}(t) \bigg( \frac{1}{P_{t} M \sigma_{h}^{2} \Bar{\beta}_{c^{\prime}}} \sum_{i = 1}^{I} y_{IS,c^{\prime}}^{i,n}(t) \bigg).
\end{align}
Substituting \eqref{eq:sin11} into \eqref{eq:ps_received}, we can write $y_{PS}^{n}$ as $y_{PS}^{n}(t) = \sum_{l = 1}^{9} y_{PS,l}^{n}$, with $\lambda_{t,c} = \frac{P_{IS,t}}{K K^{\prime} M \sigma_{h}^{2}}$, each term can be written as
\begin{align*}
    \hspace{-6.8cm} y_{PS,1}^{n}(t) \! = \! \sum_{c,m,i,k,k^{\prime}} \frac{\lambda_{t,c}}{\Bar{\beta}_{c}} | h_{PS,c,k^{\prime}}^{n}(t) |^{2}  |h_{c,m,c,k}^{i,n}(t)|^{2} \Delta \theta_{c,m}^{i,n,cx}(t),
\end{align*}
\begin{align*}
    \hspace{-4cm} y_{PS,2}^{n}(t) \! = \! \sum_{c,m,m^{\prime} \neq m,i,k,k^{\prime}} \frac{\lambda_{t,c}}{\Bar{\beta}_{c}} | h_{PS,c,k^{\prime}}^{n}(t) |^{2} \big( h_{c,m,c,k}^{i,n}(t) \big)^{\ast} h_{c,m^{\prime},c,k}^{i,n}(t) \Delta \theta_{c,m^{\prime}}^{i,n,cx}(t),
\end{align*}
\begin{align*}
    \hspace{-3.9cm} y_{PS,3}^{n}(t) \! =  \! \sum_{c,c^{\prime} \neq c, m, m^{\prime},i,k,k^{\prime}} \frac{\lambda_{t,c}}{\Bar{\beta}_{c}}  | h_{PS,c,k^{\prime}}^{n}(t) |^{2} \! \big( \! h_{c,m,c,k}^{i,n}(t) \! \big)^{\ast} h_{c,m^{\prime},c^{\prime},k}^{i,n}(t) \Delta \theta_{c^{\prime},m^{\prime}}^{i,n,cx}(t),
\end{align*}
\begin{align*}
    \hspace{-6.3cm} y_{PS,4}^{n}(t) \! = \! \sum_{c, m,i,k,k^{\prime}} \frac{\lambda_{t,c}}{P_{IS,t} \Bar{\beta}_{c}} |h_{PS,c,k^{\prime}}^{n}(t) |^{2} \big( h_{c,m,c,k}^{i,n}(t) \big)^{\ast} z_{IS,c,k}^{i,n}(t), 
\end{align*}
\begin{align*}
    \hspace{-4cm} y_{PS,5}^{n}(t) \! = \! \sum_{c,c^{\prime} \neq c, m,i,k,k^{\prime}} \frac{\lambda_{t,c}}{\Bar{\beta}_{c^{\prime}}} \big( h_{PS,c,k^{\prime}}^{n}(t) \big)^{\ast} h_{PS,c^{\prime},k^{\prime}}^{n}(t) |h_{c^{\prime},m,c^{\prime},k}^{i,n}(t)|^{2}  \Delta \theta_{c^{\prime},m}^{i,n,cx}(t),  
\end{align*}
\begin{align*}
    \hspace{-1.3cm} y_{PS,6}^{n}(t) \! = \! \sum_{c,c^{\prime} \neq c, m, m^{\prime} \neq m,i,k,k^{\prime}} \frac{\lambda_{t,c}}{\Bar{\beta}_{c^{\prime}}} \! \left( \! h_{PS,c,k^{\prime}}^{n}(t) \! \right)^{\ast} \! h_{PS,c^{\prime},k^{\prime}}^{n}(t) \! \left( \! h_{c^{\prime},m,c^{\prime},k}^{i,n}(t) \! \right)^{\ast} \! h_{c^{\prime},m^{\prime},c^{\prime},k}^{i,n}(t) \Delta \theta_{c^{\prime},m^{\prime}}^{i,n,cx}(t),
\end{align*}
\begin{align*}
    \hspace{-0.3cm} y_{PS,7}^{n}(t) \! = \! \sum_{c,c^{\prime} \neq c, c^{\prime \prime} \neq c^{\prime}, m, m^{\prime},i,k,k^{\prime}} \frac{\lambda_{t,c}}{\Bar{\beta}_{c^{\prime}}} \left( h_{PS,c,k^{\prime}}^{n}(t) \right)^{\ast} h_{PS,c^{\prime},k^{\prime}}^{n}(t) \left( h_{c^{\prime},m,c^{\prime},k}^{i,n}(t) \right)^{\ast} h_{c^{\prime},m^{\prime},c^{\prime \prime},k}^{i,n}(t) \Delta \theta_{c^{\prime \prime},m^{\prime}}^{i,n,cx}(t),   
\end{align*}
\begin{align*}
    \hspace{-3.3cm} y_{PS,8}^{n}(t) \! = \! \sum_{c,c^{\prime} \neq c, m,i,k,k^{\prime}} \frac{\lambda_{t,c}}{P_{IS,t} \Bar{\beta}_{c^{\prime}}} \left( h_{PS,c,k^{\prime}}^{n}(t) \right)^{\ast}  h_{PS,c^{\prime},k^{\prime}}^{n}(t) \big( h_{c^{\prime},m,c^{\prime},k}^{i,n}(t) \big)^{\ast} z_{IS,c^{\prime},k}^{i,n}(t), 
\end{align*}
\begin{align}
    \hspace{-8.4cm} y_{PS,9}^{n}(t) \! = \! \sum_{c,k^{\prime}} \frac{1}{K^{\prime}} \big( h_{PS,c,k^{\prime}}^{n}(t) \big)^{\ast} z_{PS,k}^{n}(t), \label{eq: convterms}
\end{align}
\section{Convergence Analysis} \label{sec:convanalysis}
In this section, we present an upper bound on the global loss function, which shows how far the global FL model is after a certain number of iterations from the optimal model. Define the optimal solution that minimizes the loss $F(\bm{\theta})$ as
\begin{equation}
    \bm{\theta}^{\ast} \triangleq \argmin_{\bm{\theta}} F(\bm{\theta}).
\end{equation}
Also, the minimum value of the loss function is denoted as $F^{*} = F(\bm{\theta}^{*})$, the minimum value of the local loss function $F_{c,m}$ is given as $F_{c,m}^{*}$, and the bias in the dataset is defined as
\begin{equation}
    \Gamma \triangleq F^{*} - \sum_{c = 1}^{C} \sum_{m = 1}^{M} \frac{B_{c,m}}{B} F_{c,m}^{*} \geq 0.
\end{equation}
In addition, we assume that the learning rate of the overall system does not change in user and cluster iterations, i.e., $\eta_{c,m}^{i,j}(t) = \eta(t)$. Therefore, we can write the global update rule as
\begin{align}
    \bm{\theta}_{c,m}^{i,j+1}(t) & = \bm{\theta}_{c,m}^{i,j}(t) - \eta(t) \nabla F_{c,m} (\bm{\theta}_{c,m}^{i,j}(t), \bm{\xi}_{c,m}^{i,j}(t)),
\end{align}
which can also be written as
\begin{equation}
    \bm{\theta}_{c,m}^{i,j+1}(t) - \bm{\theta}_{c,m}^{i,1}(t) = - \eta(t) \sum_{l = 1}^{j} \nabla F_{c,m} (\bm{\theta}_{c,m}^{i,l}, \bm{\xi}_{c,m}^{i,l}(t)).
    \label{eq: useriteration}
\end{equation}
We make the following two assumptions as in \cite{amiri2021blind}.
\begin{assumption} \label{assumption1}
All the loss functions $F_{1,1}, \ldots, F_{C,M}$ for all the clusters and users are L-smooth and $\mu$-strongly convex, which are, respectively $\forall \bm{v},\bm{w} \in \mathbb{R}^{2N}$, $\forall m \in [M], \forall c \in [C]$,
\begin{align}
    \!\!F_{c,m}(\bm{v}) \!-\! F_{c,m}(\bm{w}) & \!\leq\! \langle \bm{v} \!-\! \bm{w}, \! \nabla\! F_{c,m} (\bm{w}) \rangle \!+\! \frac{L}{2} \!\left\| \bm{v} - \bm{w} \right\|_{2}^{2}\!, \\
    \!\!F_{c,m}(\bm{v}) \!-\! F_{c,m}(\bm{w}) & \!\geq\! \langle \bm{v} \!-\! \bm{w},  \!\nabla\! F_{c,m} (\bm{w}) \rangle \!+\! \frac{\mu}{2} \!\left\| \bm{v} - \bm{w} \right\|_{2}^{2}\!.
\end{align}
\end{assumption}
\begin{assumption} \label{assumption2}
The expected value of the squared $l_{2}$ norm of the stochastic gradients are bounded, which is, $\forall j \in [\tau], i \in [I], \forall m \in [M], \forall c \in [C]$, 
    $\mathbb{E}_{\xi} \left[  \left\| \nabla F_{c,m} (\bm{\theta}_{c,m}^{i,j}(t), \bm{\xi}_{c,m}^{i,j}(t)) \right\|_{2}^{2} \right] \leq G^{2}$,
which in turn translates into $\mathbb{E}_{\xi} \left[  \nabla F_{c,m}(\theta_{c,m}^{i,j,n}(t), \xi_{c,m}^{i,j,n}(t)) \right] \leq G$, $\forall n \in [2N]$.
\end{assumption}
\begin{theorem} \label{thm1} 
In W-HFL, for $0 \leq \eta(t) \leq min\Big\{ 1, \frac{1}{\mu \tau I} \Big\}$, the global loss function can be upper bounded as
\begin{align}
    \mathbb{E} \left[ \left\| \bm{\theta}_{PS}(t) - \bm{\theta}^{*} \right\|_{2}^{2} \right] \leq \! \left( \prod_{a = 0}^{t-1} \! X(a) \! \right) \! \left\| \bm{\theta}_{PS}(0) \! - \! \bm{\theta}^{*} \right\|_{2}^{2} \! + \! \sum_{b = 0}^{t-1} \! Y(b) \! \prod_{a = b+1}^{t-1} \! X(a), \label{ourt1}
\end{align}
where $X(t) = \left( 1 - \mu \eta(t) I \left( \tau - \eta(t) (\tau - 1) \right) \right)$, and
\begin{align} 
    & Y(t) = \frac{\eta^{2}(t) G^{2} I^{2} \tau^{2}}{M^{2}C^{2}} \sum_{c_{1} = 1}^{C} \sum_{c_{2} = 1}^{C} \sum_{m_{1} = 1}^{M} \sum_{m_{2} = 1}^{M} A(m_{1},m_{2},c_{1},c_{2}) \nonumber \\
    & + \frac{\big( 2 \! + \! (M\!\!-\!\!1)(C\!\!-\!\!2)(K\!\!-\!\!1)(I\!\!-\!\!1) \big) \eta^{2}(t) I G^{2} \tau^{2}}{K (K^{\prime}) M^3 C^2 (C-1) \Bar{\beta}^{2}} \!\! \sum_{c = 1}^{C} \sum_{\substack{c^{\prime} = 1 \\ c^{\prime} \neq c}}^{C} \! \sum_{m_{1} = 1}^{M} \sum_{m_{2} = 1}^{M} \frac{\beta_{IS,c} \beta_{IS,c^{\prime}} \beta_{c^{\prime},m_{1},c^{\prime}} \beta_{c^{\prime},m_{2},c^{\prime}}}{\Bar{\beta}_{c^{\prime}}^{2}} \nonumber \\
    & + \frac{\eta^{2}(t) G^{2} I \tau^{2} }{K K^{\prime}M^{2}C^{2} \Bar{\beta}^{2}} \sum_{c = 1}^{C} \sum_{m = 1}^{M} \Bigg( \frac{(K^{\prime}+1) \beta_{IS,c}^{2} \beta_{c,m,c}}{\Bar{\beta}_{c}^{2}} \bigg( \sum_{\substack{m^{\prime} = 1 \\ m^{\prime} \neq m}}^{M} \beta_{c,m^{\prime},c} + \sum_{\substack{c^{\prime} = 1 \\ c^{\prime} \neq c}}^{C} \sum_{m^{\prime} = 1}^{M} \beta_{c,m^{\prime},c^{\prime}} \bigg) \Bigg) \nonumber \\
    & + \frac{\eta^{2}(t) G^{2} I \tau^{2} }{K K^{\prime}M^{2}C^{2} \Bar{\beta}^{2}} \sum_{c = 1}^{C} \sum_{\substack{c^{\prime} = 1 \\ c^{\prime} \neq c}}^{C} \sum_{m = 1}^{M} \Bigg( \frac{\beta_{IS,c} \beta_{IS,c^{\prime}} \beta_{c^{\prime},m,c^{\prime}}}{\Bar{\beta}_{c^{\prime}}^{2}} \bigg( \sum_{\substack{m^{\prime} = 1 \\ m^{\prime} \neq m}}^{M} \beta_{c^{\prime},m^{\prime},c^{\prime}} + \sum_{\substack{c^{\prime \prime} = 1 \\ c^{\prime \prime} \neq c^{\prime}}}^{C} \beta_{c^{\prime},m^{\prime},c^{\prime \prime}} \bigg) \Bigg) \nonumber \\
    & + \frac{\sigma_{z}^{2} N }{K^{\prime} C^{2} \sigma_{h}^{2} \Bar{\beta}^{2}} \sum_{c = 1}^{C} \beta_{IS,c} \Bigg( \frac{1}{P_{IS,t}^{2}} + \frac{I}{K M^2} \sum_{m = 1}^{M} \bigg( \frac{(K^{\prime}+1) \beta_{IS,c} \beta_{c,m,c}}{P_{t}^{2} \Bar{\beta}_{c}^{2}} + \sum_{\substack{c^{\prime} = 1 \\ c^{\prime} \neq c}}^{C} \frac{\beta_{IS,c^{\prime}} \beta_{c^{\prime},m,c^{\prime}}}{P_{IS,t}^{2} \Bar{\beta}_{c^{\prime}}^{2}} \bigg) \Bigg) \nonumber \\
    & + \! \left( 1 \! + \! \mu (1 \! - \! \eta(t) \right) \eta^{2}(t) I G^{2} \frac{\tau (\tau \! - \! 1) (2 \tau \! - \! 1)}{6} + \eta^{2}(t) I (\tau^{2} \! + \! \tau \! - \! 1) G^{2} + 2 \eta(t) I (\tau \! - \! 1) \Gamma, \label{eq:xtyt}
\end{align}
with $A(m_{1},m_{2},c_{1},c_{2}) \! = \! 1  -  \mathlarger{\frac{\beta_{c_{1},m_{1},c_{1}} \beta_{IS,c_{1}}}{ \Bar{\beta} \Bar{\beta_{c_{1}}} }  -  \frac{\beta_{c_{2},m_{2},c_{2}} \beta_{IS,c_{2}}}{ \Bar{\beta} \Bar{\beta_{c_{2}}} } + \frac{ \beta_{c_{1},m_{1},c_{1}} \beta_{c_{2},m_{2},c_{2}} \beta_{IS,c_{1}} \beta_{IS,c_{2}}}{M C K K^{\prime} I \Bar{\beta}^{2} \Bar{\beta}_{c_{1}} \Bar{\beta}_{c_{2}}}} \\ \times \big( 4 \! + \! 2(K^{\prime} \! - \!1) \! + \! (M \!- \! 1)(K \! - \! 1)(I \! - \! 1)\big( 2 \! + \! (K^{\prime} \! - \! 1)(C \! - \! 1) \big) \big)$.
\end{theorem}
\begin{IEEEproof} 
See Appendix A.
\end{IEEEproof}
\begin{corollary} \label{cor1}
Assuming L-smoothness, after $T$ global iterations, the loss function can be upper-bounded as
\begin{align}
    \mathbb{E} \left[ F \big( \bm{\theta}_{PS}(T) \big) \right] - F^{*} & \leq \frac{L}{2} \mathbb{E} \left[ \left\| \bm{\theta}_{PS}(T) - \bm{\theta}^{*} \right\|_{2}^{2} \right], \! \nonumber \\
    & \leq \frac{L}{2} \left( \prod_{a = 0}^{T-1} \! X(a) \! \right) \! \left\| \bm{\theta}_{PS}(0) \! - \! \bm{\theta}^{*} \right\|_{2}^{2} \! + \! \frac{L}{2} \sum_{b = 0}^{T-1} \! Y(b) \! \prod_{a = b+1}^{T-1} \! X(a). \label{ourc1}
\end{align}
\end{corollary}
\noindent {\textbf{Remark 1.}} Since the fourth term in $Y(a)$ is independent of $\eta(a)$, even for $\Lim{t \to \infty} \eta(t) = 0$, we have $\Lim{t \to \infty} \mathbb{E} [F(\bm{\theta}_{PS}(t))] - F^{*} \neq 0$. $Y(a)$ is also proportional to $I$ and $\tau$, meaning that more user iterations and cluster aggregations do not always provide faster convergence. However, since the MUs experience lower path-loss in W-HFL than in the conventional FL, it can reach a higher accuracy. Moreover, increasing the number of clusters $C$ leads to a faster convergence, however, at the cost of employing more ISs. 
\begin{corollary}
For a simplified setting with $I = \tau = 1, P_{IS,t} \gg P_{t}, P_{t} = P, \forall t, \beta_{c,m,c} = \beta, \beta_{IS,c} = \beta_{IS}, \forall m \in [M], \forall c \in [C]$, we have $X(t) = \big( 1 - \mu \eta(t) \big)$ and 
\begin{align}
    Y(t) & \approx \frac{\eta^{2}(t)G^{2}}{K K^{\prime} M^{3} C^{3}} M C K^{\prime} + 2\eta^{2}G^{2} \bigg( 1 - \frac{1}{M C} \bigg) + \frac{\sigma_{z}^{2} N}{K C^{3} \sigma_{h}^{2}} \bigg( \frac{1}{P_{IS,t}} + \frac{(K^{\prime} + 1)}{K M^{3} P^{2}} + \frac{(C-1)}{P_{IS,t}^{2} M^{2}} \bigg) \nonumber \\
    & \approx \frac{\eta^{2}(t)G^{2}}{K M^{2} C^{2}} + 2 \eta^{2} G^{2} + \frac{\sigma_{z}^{2} N}{K M^{3} C^{3} \sigma_{h}^{2} P^{2}} \nonumber \\
    & \approx 2 \eta^{2}(t) G^{2} + \frac{\sigma_{z}^{2} N}{K M^{3} C^{3} \sigma_{h}^{2} P^{2}},
\end{align}
which, when $\eta(t) = \eta, \forall t$, simplify the upper bound on the loss function as
\begin{align}
    \mathbb{E} \left[ F \big( \bm{\theta}_{PS}(T) \big) \right] - F^{*} & \leq \frac{L}{2} \big( 1 - \mu \eta \big)^{T} \! \left\| \bm{\theta}_{PS}(0) \! - \! \bm{\theta}^{*} \right\|_{2}^{2} \nonumber \\
    & \hspace{0.5cm} + \! \frac{L}{2 \mu \eta} \Big( 2 \eta^{2} G^{2} + \frac{\sigma_{z}^{2} N}{K M^{3} C^{3} \sigma_{h}^{2} P^{2}} \Big) \Big( 1 - \big( 1 - \mu \eta \big)^{T} \Big). \label{ourc22}
\end{align}
\end{corollary}
\noindent {\textbf{Remark 2.}} As expected, it can be observed that the numbers of receive antennas, MUs and ISs have a positive effect on the convergence, whereas the model dimension has an adverse effect.
\section{Simulation Results} \label{sec:simresults}
In this section, we evaluate and compare the performance of W-HFL with that of the conventional FL under different scenarios. Via different experiments, we observe the power consumption, as well as the convergence speed of the learning algorithm with different number of cluster aggregations, $I$. In our experiments, we use two different image classification datasets, MNIST \cite{mnist} and CIFAR-10 \cite{cifar10}. For the MNIST dataset, we train a single layer neural network with 784 input neurons and 10 output neurons with $2N = 7850$; and, for CIFAR-10, we employ a convolutional neural network (CNN) architecture which has two $3 \times 3 \times 32$, two $3 \times 3 \times 64$, and two $3 \times 3 \times 128$ convolutional layers, each of them with the same padding, batch normalization, and ReLU function. After every two convolutional layers, it has $2 \times 2$ max pooling and dropout with $0,2, 0.3,$ and $0.4$, respectively. In the end, we have a fully-connected layer with 10 output neurons and softmax activation, which corresponds to $2N = 307498$. We employ Adam optimizer \cite{adam} for training both networks.

We consider a hierarchical structure with $D = 20$ MUs, $C = 4$ circular clusters each with a single IS in the middle and $M = 5$ MUs in each cluster, and a single PS. MUs in each cluster are randomly distributed at a normalized distance between 0.5 and 1 units from their corresponding IS. Also, these clusters are randomly placed at a normalized distance between 0.5 and 3 units from the PS. 

The experiments are performed with two different data distributions. In the i.i.d. experiments, all the training data is randomly and equally distributed across MUs. In the non-i.i.d. case, we split the training data into $3MC$ groups each consisting of samples with the same label, and randomly assign 3 groups to each MU randomly. As a second non-i.i.d. case, we distribute the labels to different clusters in such a way that each cluster pair has 6 shared labels, and assigned labels are distributed randomly across MUs in each cluster. In order to make the comparison fair, we use a normalized time $IT$ in the accuracy plots where $T$ denotes the number of global iterations. 

In the experiments, the total time $IT$ is set to 400, where it is assumed that the conventional FL has $I = 1$, the mini-batch size is $|\bm{\xi}_{c,m}^{i}(t)| = 500$ for MNIST training and $|\bm{\xi}_{c,m}^{i}(t)| = 128$ for CIFAR-10 training, the path loss exponent $p$ is set to 4,  $\sigma_{h}^{2} = 1$, $\sigma_{z}^{2} = 10$ for the MNIST, and $\sigma_{z}^{2} = 1$ for the CIFAR-10 training. Each IS and the PS has $5MC = 100$ receive antennas. Also, the power multipliers are set to $P_{t} = 1 + 10^{-2} t$, and $P_{IS,t} = 20P_{t}$, $t \in [T]$. In order to make the average transmit power levels consistent among different simulations, $P_{t,low} = 0.5 P_{t}$ is used for the cases with $I = 1$.
\begin{figure}
     \centering
     \begin{subfigure}[b]{0.32\textwidth}
         \centering
         \includegraphics[width=\textwidth]{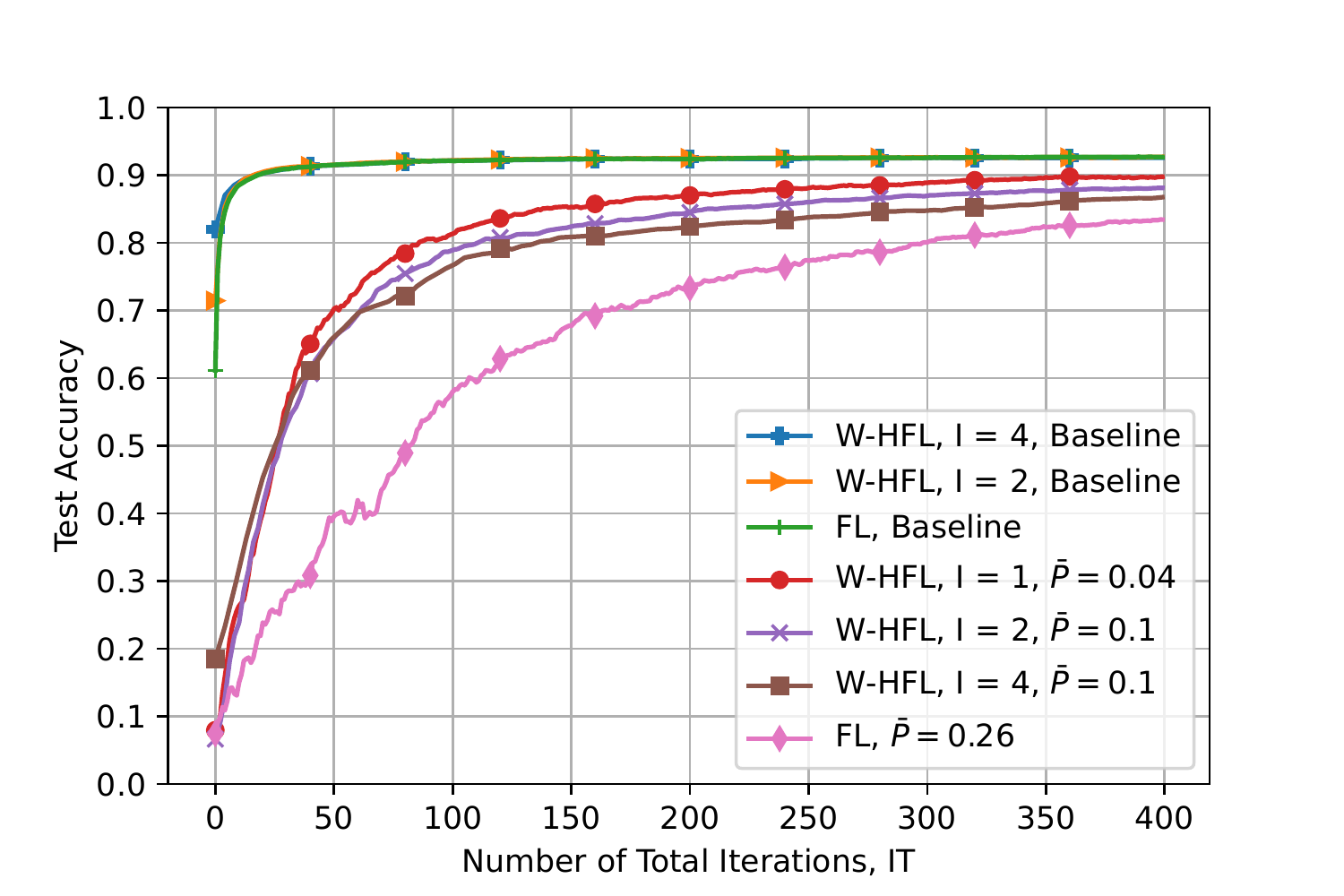}
         \caption{I.i.d. with $\tau = 1$.}
         \label{fig:iid_mnist}
     \end{subfigure}
     \hfill
     \begin{subfigure}[b]{0.32\textwidth}
         \centering
         \includegraphics[width=\textwidth]{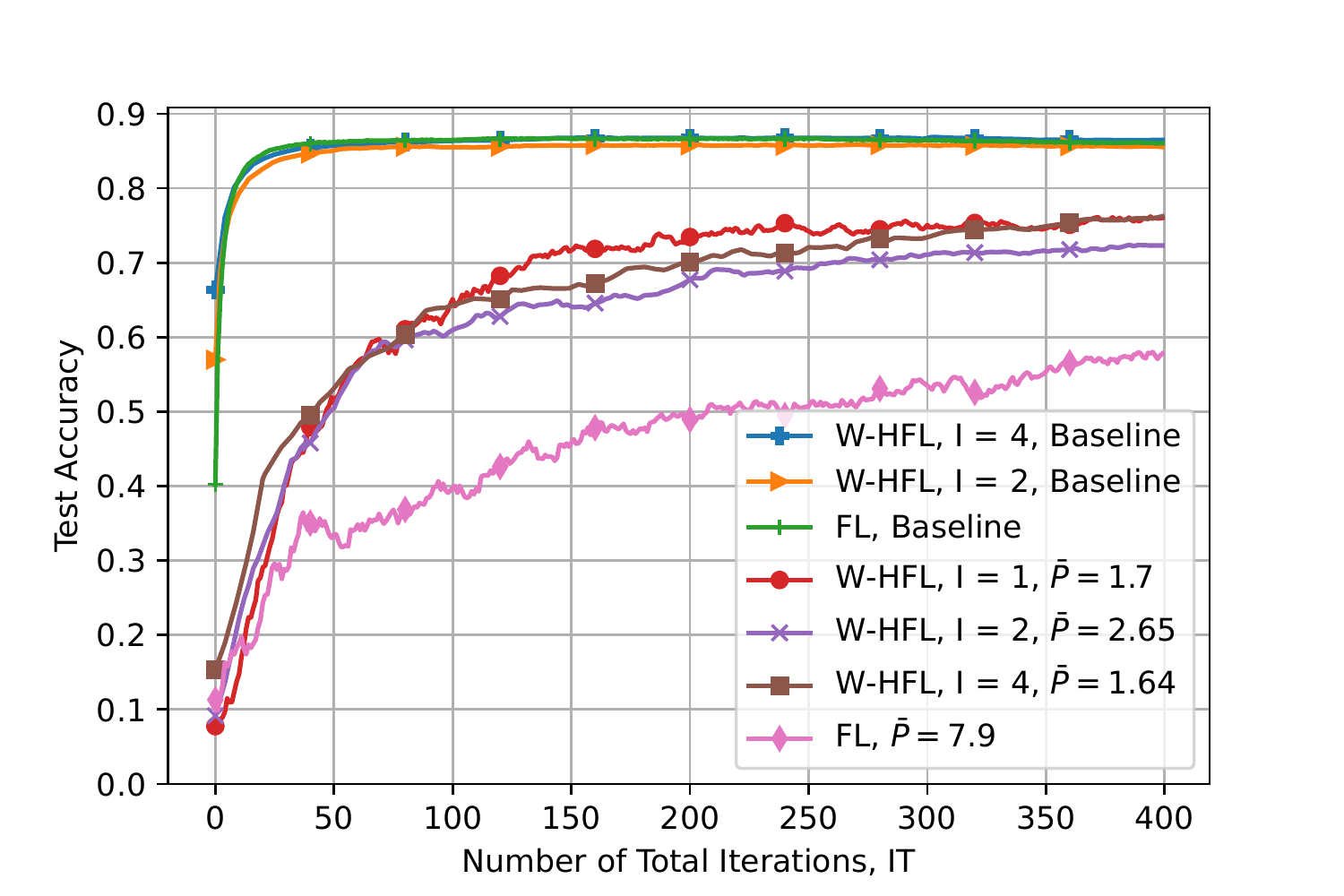}
         \caption{Non-i.i.d. MUs with $\tau = 3$.}
         \label{fig:noniid_mnist}
     \end{subfigure}
      \begin{subfigure}[b]{0.32\textwidth}
     \centering
     \includegraphics[width=\textwidth]{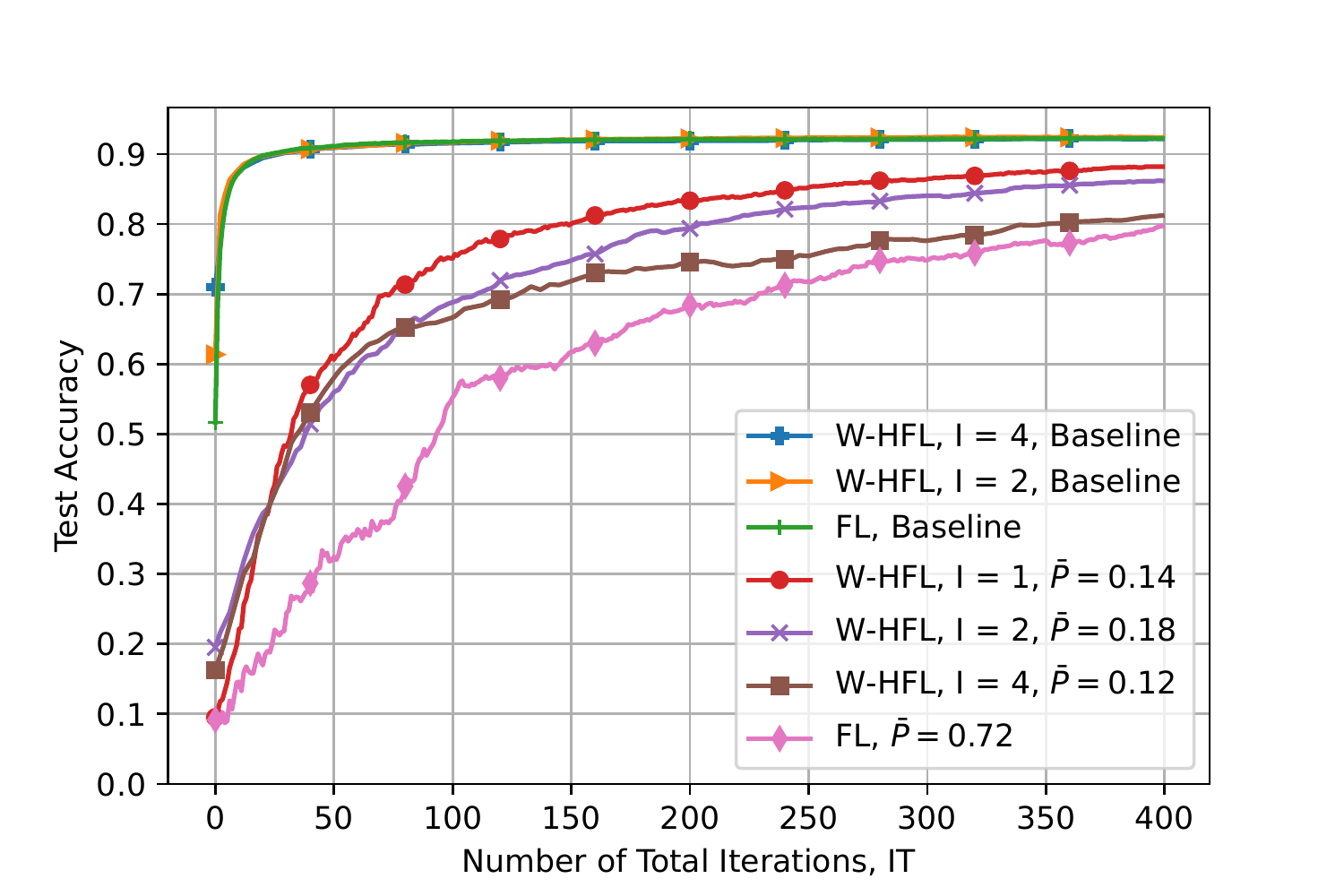}
     \caption{Non-i.i.d. clusters with $\tau = 1$.}
     \label{fig:clusternoniid_mnist}
 \end{subfigure}
    \caption{Test accuracy for the MNIST dataset.}
    \label{fig:mnist}
\end{figure}

In Fig. \ref{fig:mnist}, we present the performance of W-HFL with different number of cluster aggregations $I$ using the MNIST dataset. We also report the average transmit power per total number of iterations at the edge for each case. We consider W-HFL with $I = 1$, $I = 2$, and $I = 4$, as well as the conventional FL scheme, i.e., $I = 1$, with no IS in-between. To assess their performance, we compare the results with ideal baseline cases, where the model differences are assumed to be transmitted in an error-free manner. We can observe in Fig. \ref{fig:iid_mnist} that W-HFL outperforms conventional FL while using less power at the edge. This is mainly because in W-HFL, MUs have a closer server (IS) to transmit their signals to, thereby being less affected by the path-loss effects. Also, it can be seen that the performance slightly deteriorates as $I$ increases, while consuming less transmit power at the edge. The system performs better in i.i.d. distribution when the ISs perform less cluster aggregations, and the best performance is observed with $I=1$, where the ISs just relay the received cluster updates. In Fig. \ref{fig:noniid_mnist}, we consider MNIST with non-i.i.d. distribution across MUs, and evaluate the system performances for $\tau = 3$. We can see the change in the order of performance when the distribution changes and $\tau$ increases since having more cluster updates before the global aggregation provides a more powerful update for the model than having a frequent global model update with less trained non-i.i.d. datasets. Moreover, we evaluate the performance when clusters are non-i.i.d. in Fig. \ref{fig:clusternoniid_mnist}. When the clusters are non-i.i.d, we observe a slight decrease in accuracies when compared to the i.i.d. data distribution.
\begin{figure}
    \centering
    \begin{minipage}{0.5\textwidth}
         \centering
         \includegraphics[width=1\textwidth]{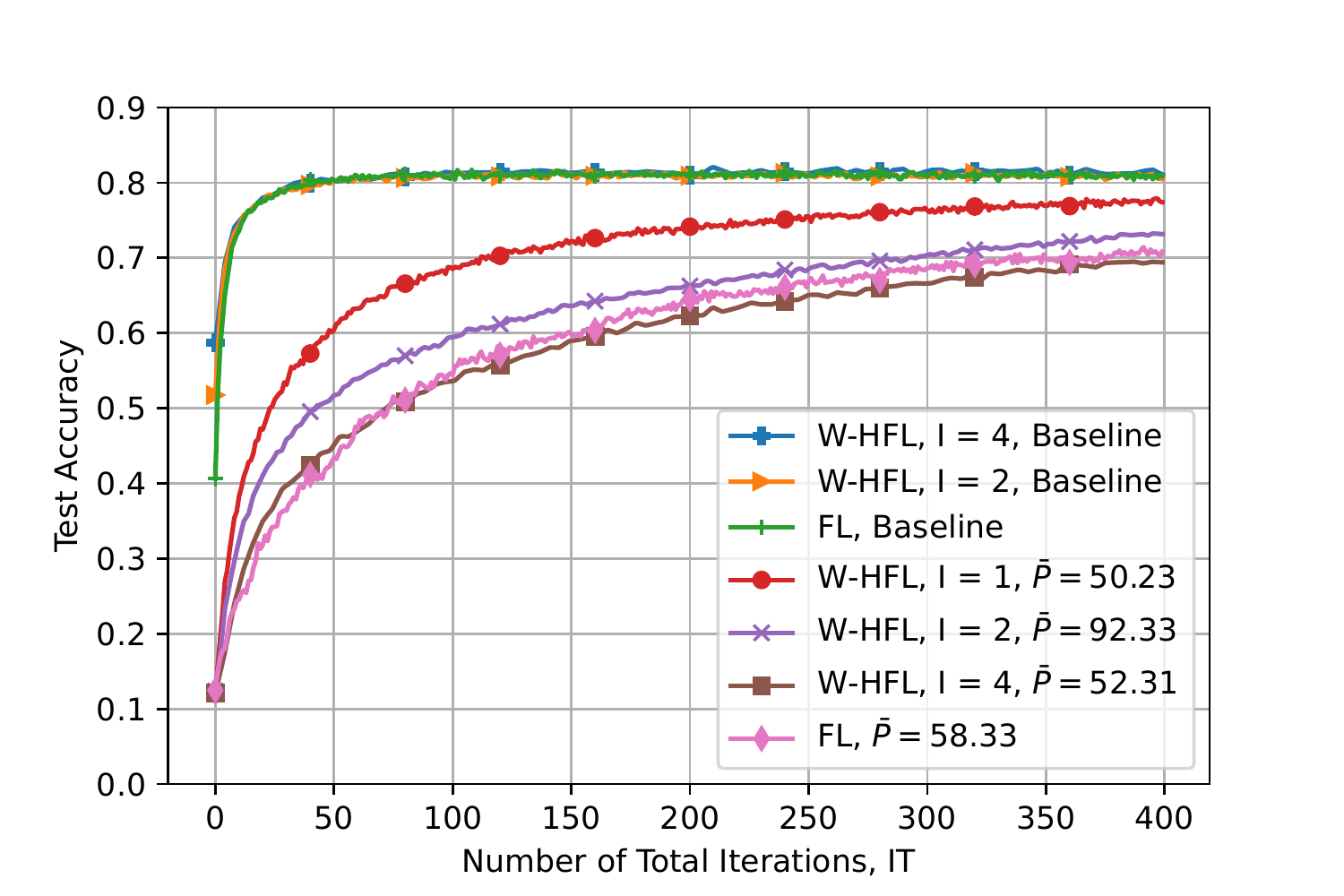}
         \caption{CIFAR-10 training with $\tau = 5$.}
         \label{fig:iid_cifar}
    \end{minipage}%
    \begin{minipage}{0.5\textwidth}
         \centering
         \includegraphics[width=1\textwidth]{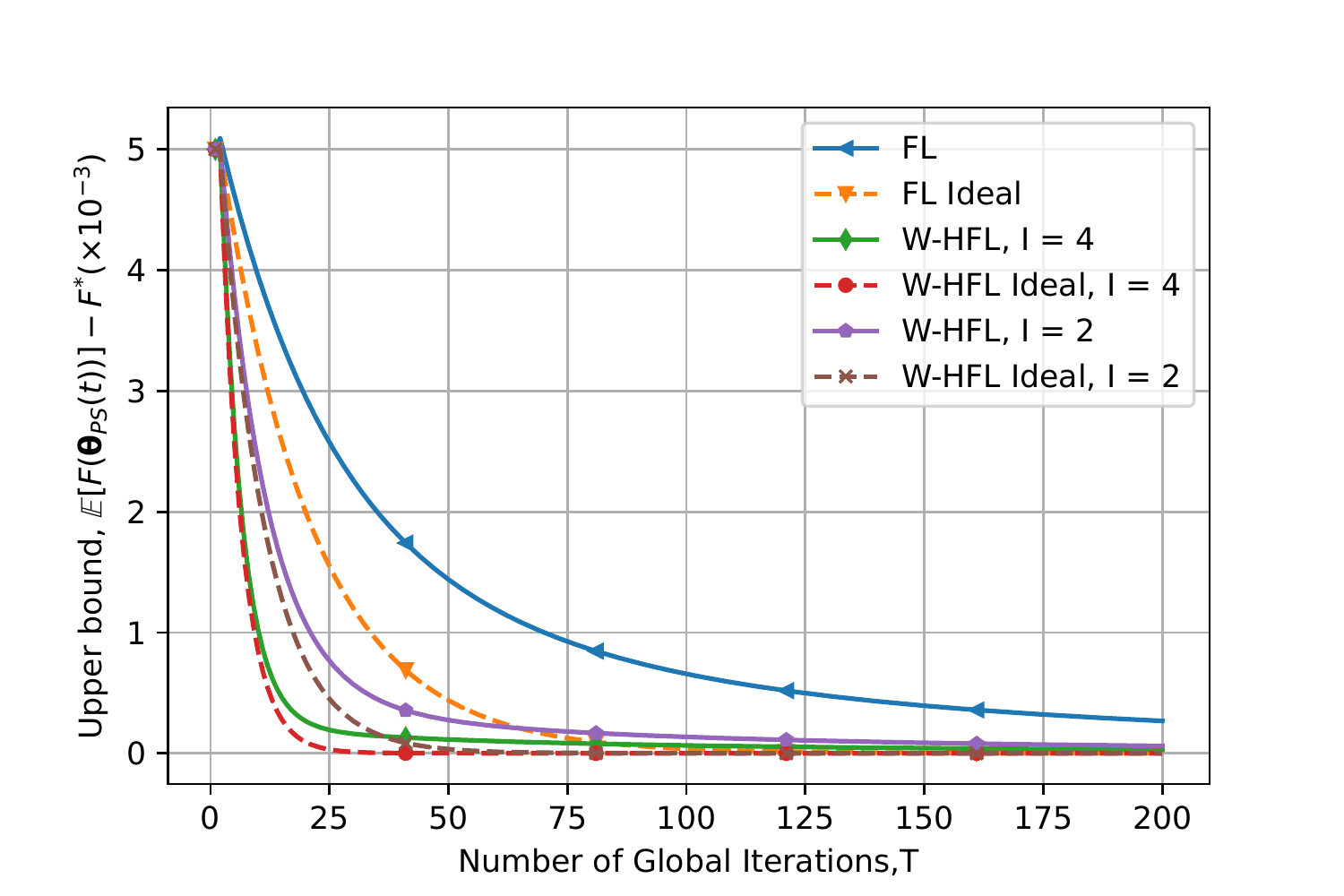}
         \caption{Convergence rate for Fig. \ref{fig:iid_mnist}.}
         \label{fig:upper_bound}
    \end{minipage}
\end{figure}

In Fig. \ref{fig:iid_cifar}, we also depict the performance of the proposed algorithm on the CIFAR-10 dataset with i.i.d. data distribution across MUs. We can see a similar trend with the i.i.d. MNIST results. However, the average transmit power values have increased when compared to MNIST results since the used model contains more parameters in CIFAR-10 simulations to tackle with the more challenging dataset. It can be observed that using the ISs as relays gives the best performance while using less transmit power. W-HFL with $I=2$ uses the more transmit power than $I=4$ since it performs more global iteration rounds with an increased $P_{t}$. We can also see that the gap between conventional FL and W-HFL is closed, and the main reason is that the transmit power is a lot higher than the noise variance since the more challenging datasets are more susceptible to wireless channel effects.

In Fig. \ref{fig:upper_bound}, we numerically analyze the convergence rate of W-HFL, with the results presented in Corollary \ref{cor1}. The setting from MNIST i.i.d. training is used with $2N \!=\! 7850, L \!=\! 10, \mu \!=\! 1, G^{2} \!=\! 1, \Gamma \!=\! 1, \eta(t) \!=\! 5 \! \cdot \! 10^{-2} \! - \! 2 \! \cdot \!  10^{-5}t, P_{t} \!=\! 1 \!+\! 10^{-2}t, P_{IS,t} \! = \! 10 P_{t}, \left\| \bm{\theta}_{PS}(0) \!-\! \bm{\theta}^{*} \right\|_{2}^{2} \!=\! 10^{3}$. We can observe that W-HFL convergences faster than the conventional FL, and performs similar to the baseline.
\section{Conclusions} \label{sec:conclusion}
We have proposed a W-HFL scheme, where edge devices exploit nearby local servers called ISs for model aggregation. After several OTA cluster aggregations, ISs transmit their model differences to the PS to update global model for the next iteration. We have considered the inter-cluster interference at the cluster aggregations, and introduced OTA aggregation also at the PS. We provided a detailed system model as well as a convergence analysis for the proposed algorithm that gives an upper bound on the global loss function. We showed through numerical and experimental analyses with different data distributions and datasets that bringing the server-side closer to the more densely located MUs can improve the final model accuracy and result in faster convergence compared to the conventional FL. We also observed that using less cluster aggregations in W-HFL can lead to higher accuracies, but with an increased cost of transmit power at the edge. 
%
\appendices
\section*{Appendix A: Proof of Theorem \ref{thm1}} 
Let us define auxiliary variable
   $ \bm{v}(t+1) = \bm{\theta}_{PS}(t) + \Delta \bm{\theta}_{PS}(t)$.
Then, we have
\begin{align}
    &\!\!\big\| \bm{\theta}_{PS}(t\!+\!1) \!-\! \bm{\theta}^{\ast} \big\|_{2}^{2} \nonumber 
    \!=\! \big\| \bm{\theta}_{PS}(t\!+\!1) \!-\! \bm{v}(t\!+\!1) + \bm{v}(t\!+\!1) \!-\! \bm{\theta}^{\ast} \big\|_{2}^{2} \nonumber \\
    &= \! \big\| \bm{\theta}_{PS}(t+1) \! - \! \bm{v}(t+1) \big\|_{2}^{2} \! + \! \big\| \bm{v}(t+1) \! - \! \bm{\theta}^{\ast} \big\|_{2}^{2} \! + \! 2 \langle \bm{\theta}_{PS}(t+1) \! - \! \bm{v}(t+1) , \bm{v}(t+1) \! - \! \bm{\theta}^{\ast} \rangle. \label{eq: theorem1argument}
\end{align}
Next, we provide upper bounds on the three terms of (\ref{eq: theorem1argument}).
\begin{lemma} \label{convproof1} 
$\mathbb{E} \Big[ \big\| \bm{\theta}_{PS}(t+1) - \bm{v}(t+1) \big\|_{2}^{2} \Big] \leq \mathlarger{ \frac{\eta^{2}(t) G^{2} I^{2} \tau^{2}}{M^{2}C^{2}} \sum_{c_{1} = 1}^{C} \sum_{c_{2} = 1}^{C} \sum_{m_{1} = 1}^{M} \sum_{m_{2} = 1}^{M}} A(m_{1},m_{2},c_{1},c_{2}) \\
    + \mathlarger{\frac{\big( 2 + (M\!-\!1)(C\!-\!2)(K\!-\!1)(I\!-\!1) \big) \eta^{2}(t) I G^{2} \tau^{2}}{K (K^{\prime}) M^3 C^2 (C-1) \Bar{\beta}^{2}} \sum_{c = 1}^{C} \sum_{\substack{c^{\prime} = 1 \\ c^{\prime} \neq c}}^{C} \sum_{m_{1} = 1}^{M} \sum_{m_{2} = 1}^{M} \frac{\beta_{IS,c} \beta_{IS,c^{\prime}} \beta_{c^{\prime},m_{1},c^{\prime}} \beta_{c^{\prime},m_{2},c^{\prime}}}{\Bar{\beta}_{c^{\prime}}^{2}}} \\
    + \mathlarger{\frac{\eta^{2}(t) G^{2} I \tau^{2} }{K K^{\prime}M^{2}C^{2} \Bar{\beta}^{2}} \sum_{c = 1}^{C} \sum_{m = 1}^{M} \Bigg( \frac{(K^{\prime}+1) \beta_{IS,c}^{2} \beta_{c,m,c}}{\Bar{\beta}_{c}^{2}} \bigg( \sum_{\substack{m^{\prime} = 1 \\ m^{\prime} \neq m}}^{M} \beta_{c,m^{\prime},c} + \sum_{\substack{c^{\prime} = 1 \\ c^{\prime} \neq c}}^{C} \sum_{m^{\prime} = 1}^{M} \beta_{c,m^{\prime},c^{\prime}} \bigg) \Bigg)} \\
    + \mathlarger{ \frac{\eta^{2}(t) G^{2} I \tau^{2} }{K K^{\prime}M^{2}C^{2} \Bar{\beta}^{2}} \sum_{c = 1}^{C} \sum_{\substack{c^{\prime} = 1 \\ c^{\prime} \neq c}}^{C} \sum_{m = 1}^{M} \Bigg( \frac{\beta_{IS,c} \beta_{IS,c^{\prime}} \beta_{c^{\prime},m,c^{\prime}}}{\Bar{\beta}_{c^{\prime}}^{2}} \bigg( \sum_{\substack{m^{\prime} = 1 \\ m^{\prime} \neq m}}^{M} \beta_{c^{\prime},m^{\prime},c^{\prime}} + \sum_{\substack{c^{\prime \prime} = 1 \\ c^{\prime \prime} \neq c^{\prime}}}^{C} \beta_{c^{\prime},m^{\prime},c^{\prime \prime}} \bigg) \Bigg)} \\
    + \mathlarger{ \frac{\sigma_{z}^{2} N }{K^{\prime} C^{2} \sigma_{h}^{2} \Bar{\beta}^{2}} \sum_{c = 1}^{C} \beta_{IS,c} \Bigg( \frac{1}{P_{IS,t}^{2}} + \frac{I}{K M^2} \sum_{m = 1}^{M} \bigg( \frac{(K^{\prime}+1) \beta_{IS,c} \beta_{c,m,c}}{P_{t}^{2} \Bar{\beta}_{c}^{2}} + \sum_{\substack{c^{\prime} = 1 \\ c^{\prime} \neq c}}^{C} \frac{\beta_{IS,c^{\prime}} \beta_{c^{\prime},m,c^{\prime}}}{P_{IS,t}^{2} \Bar{\beta}_{c^{\prime}}^{2}} \bigg) \Bigg)},$

where $A(m_{1},m_{2},c_{1},c_{2})$ is given in Theorem \ref{thm1}.
\end{lemma}
\begin{IEEEproof} 
See Appendix B.
\end{IEEEproof}
\begin{lemma} \label{convproof2}
$\!\mathbb{E} \Big[ \big\| \bm{v}(t + 1) \!-\! \bm{\theta}^{*} \big\|_{2}^{2} \Big] \leq \left( 1 \!-\! \mu \eta(t) I \left( \tau \!-\! \eta(t) (\tau \!-\! 1) \right) \right) \mathbb{E} \Big[ \big\| \bm{\theta}_{PS}(t) \!-\! \bm{\theta}^{*}  \big\|_{2}^{2} \Big] \\ 
\qquad + \left( 1 + \mu (1- \eta(t) \right) \eta^{2}(t) I G^{2} \mathlarger{\frac{\tau (\tau - 1) (2 \tau - 1)}{6}} + \eta^{2}(t) I (\tau^{2} + \tau - 1) G^{2} + 2 \eta(t) I (\tau - 1) \Gamma$.
\end{lemma}
\begin{IEEEproof}
See Appendix C.
\end{IEEEproof}
\begin{lemma} \label{convproof3}
$\mathbb{E} \left[ \langle \bm{\theta}_{PS}(t+1) - \bm{v}(t+1) , \bm{v}(t+1) - \bm{\theta}^{\ast}  \rangle \right] = 0$.
\end{lemma}
\begin{IEEEproof}
    $\mathbb{E} \!\left[ \langle \bm{\theta}_{\!PS}(t\!+\!1) \!\!-\!\! \bm{v}(t\!+\!1) , \bm{v}(t\!+\!1) \!\! - \!\! \bm{\theta}^{\ast}  \rangle \right] \!= \! \mathbb{E} \! \left[ \! \langle \Delta \bm{\hat{\theta}}_{\!PS}(t) \!\! - \!\! \Delta \bm{\theta}_{\!PS}(t), \! \bm{\theta}_{\!PS}(t) \! + \! \Delta \bm{\theta}_{\!PS}(t) \!\! - \!\! \bm{\theta}^{*} \rangle \! \right]$.
Then, knowing that channel realizations are independent at different user and cluster updates of the same global iteration $t$, we have
$\mathbb{E} \! \left[ \! \langle \Delta \bm{\hat{\theta}}_{\!PS}(t) \! - \! \Delta \bm{\theta}_{\!PS}(t), \! \bm{\theta}_{\!PS}(t) \! + \! \Delta \bm{\theta}_{\!PS}(t) \! - \! \bm{\theta}^{\ast} \rangle \! \right] \! = \! 0$.
\end{IEEEproof}
Recursively iterating through the results of Lemmas \ref{convproof1}, \ref{convproof2}, and \ref{convproof3} concludes the theorem.
\section*{Appendix B: Proof of Lemma \ref{convproof1}}
Using \eqref{eq: theorem1argument}, we have
\begin{align}
    \mathbb{E} \left[ || \bm{\theta}_{PS}(t+1) - \bm{v}(t+1) ||_{2}^{2} \right] &= \mathbb{E} \left[ || \Delta \hat{\bm{\theta}}_{PS}(t)  - \Delta \bm{\theta}_{PS}(t) ||_{2}^{2} \right], \\
    &= \sum_{n = 1}^{2N} \mathbb{E} \left[ ( \Delta \hat{\theta}_{PS}^{n}(t)  - \Delta \theta_{PS}^{n}(t) )^{2} \right].
\end{align}
Note that $ \Delta \hat{\theta}_{PS}^{n}(t) = \sum_{l = 1}^{9} \Delta \hat{\theta}_{PS,l}^{n}(t)$.
Using the independence of different channel realizations over different users, clusters, and the noise, we can write
\begin{equation}
    \mathbb{E} \left[ || \Delta \hat{\theta}_{PS}^{n}(t)  - \Delta \theta_{PS}^{n}(t) ||_{2}^{2} \right] = \mathbb{E} \left[ \left( \Delta \hat{\theta}_{PS,1}^{n}(t)  - \Delta \theta_{PS}^{n}(t) \right)^{2} \right] + \sum_{l = 2}^{9} \mathbb{E} \left[ \left( \Delta \hat{\theta}_{PS,l}^{n}(t) \right)^{2} \right].
\end{equation}
\begin{lemma} \label{intermediary_result1_lemma}
$\mathbb{E} \Big[ \Delta \theta_{c_{1},m_{1}}^{i_{1},n}(t) \Delta \theta_{c_{2},m_{2}}^{i_{2},n}(t) \Big] \leq \eta^{2}(t) G^{2} \tau^{2}$
\end{lemma}
\begin{IEEEproof}
$\mathbb{E} \Big[ \Delta \theta_{c_{1},m_{1}}^{i_{1},n}(t) \Delta \theta_{c_{2},m_{2}}^{i_{2},n}(t) \Big] \\
~~~~~~~~~~~~~~~~ = \! \eta^{2}(t) \mathlarger{\sum_{j_{1} = 1}^{\tau} \sum_{j_{2} = 1}^{\tau}} \mathbb{E} \bigg[ \nabla F_{c_{1},m_{1}} (\theta_{c_{1},m_{1}}^{i_{1},j_{1},n}(t), \xi_{c_{1},m_{1}}^{i_{1},j_{1},n}(t)) \nabla F_{c_{2},m_{2}} (\theta_{c_{2},m_{2}}^{i_{2},j_{2},n}(t), \xi_{c_{2},m_{2}}^{i_{2},j_{2},n}(t)) \bigg] \\
~~~~~~~~~~~~~~~~ \overset{(a)}{\leq} \eta^{2}(t) G^{2} \tau^{2},$ 

where (a) holds due to Assumption \ref{assumption2}.
\end{IEEEproof}
\begin{lemma} \label{intermediary_result2_lemma}
$\mathbb{E} \Big[ \big\| \Delta \bm{\theta}_{c,m}^{i}(t) \big\|^{2}_{2} \Big] \leq \eta^{2}(t) G^{2} \tau^{2}$
\end{lemma}
\begin{IEEEproof}
$\mathbb{E} \Big[ \big\| \Delta \bm{\theta}_{c,m}^{i}(t) \big\|^{2}_{2} \Big] = \eta^{2}(t) \mathbb{E} \bigg[ \Big\| \sum_{j = 1}^{\tau} \nabla F_{c,m} (\bm{\theta}_{c,m}^{i,j}(t), \bm{\xi}_{c,m}^{i,j}(t)) \Big\|_{2}^{2} \bigg] \\
~~~~~~~~~~~~~~~~ \overset{(a)}{\leq} \eta^{2}(t) \tau \sum_{j = 1}^{\tau} \mathbb{E} \bigg[ \Big\| \nabla F_{c,m} (\bm{\theta}_{c,m}^{i,j}(t), \bm{\xi}_{c,m}^{i,j}(t)) \Big\|_{2}^{2} \bigg] \\
~~~~~~~~~~~~~~~~ \overset{(b)}{\leq} \eta^{2}(t) G^{2} \tau^{2},$

where (a) is obtained using the convexity of $\left\|  \right\|_{2}^{2}$ and (b) holds because of Assumption \ref{assumption2}.
\end{IEEEproof}
\begin{lemma} \label{lemma11} 
    $\mathlarger{\sum_{n = 1}^{2N}} \mathbb{E} \Big[ \big( \Delta \hat{\theta}_{PS,1}^{n}(t)  - \Delta \theta_{PS}^{n}(t) \big)^{2} \Big] \leq \mathlarger{\frac{\eta^{2}(t) G^{2} I^{2} \tau^{2}}{M^{2}C^{2}} \sum_{c_{1} = 1}^{C} \sum_{c_{2} = 1}^{C} \sum_{m_{1} = 1}^{M} \sum_{m_{2} = 1}^{M}} A(m_{1},m_{2},c_{1},c_{2})$.
\end{lemma}
\begin{IEEEproof}
Using \eqref{eq: globalmodeldiff} and \eqref{eq: convterms}, we have
\begin{align*}
    & \mathbb{E} \! \Big[ \! \big( \Delta \hat{\theta}_{PS,1}^{n}(t)\! - \!\Delta \theta_{PS}^{n}(t) \! \big)^{2} \Big] \! \nonumber \\
    & = \mathbb{E} \Bigg[ \bigg( \frac{1}{MC} \sum_{c,m,i} \Delta \theta_{c,m}^{i,n}(t) \Big( \Big( \frac{1}{K K^{\prime}\sigma_{h}^{4} \Bar{\beta} \Bar{\beta}_{c}} \sum_{k,k^{\prime}} | h_{IS,c,k^{\prime}}^{n}(t) |^{2} |h_{c,m,c,k}^{i,n}(t)|^{2} \Big) - 1 \Big) \bigg)^{2} \Bigg], \nonumber 
\end{align*}
\begin{align*}
    &= \frac{1}{M^{2}C^{2}} \mathbb{E} \Bigg[ \sum_{c_{1}, c_{2}, m_{1}, m_{2}, i_{1}, i_{2}} \!\!\!\!\!\!\!\!\! \Delta \theta_{c_{1},m_{1}}^{i_{1},n}(t) \Delta \theta_{c_{2},m_{2}}^{i_{2},n}(t) \bigg( \! 1 \! - \! \frac{2}{K K^{\prime}\sigma_{h}^{4} \Bar{\beta} \Bar{\beta}_{c_{1}}} \sum_{k_{1},k_{1}^{\prime}} |h_{IS,c_{1},k^{\prime}_{1}}^{n}(t)|^{2} |h_{c_{1},m_{1},c_{1},k_{1}}^{i_{1},n}(t)|^{2} \bigg. \Bigg. \nonumber \\ 
    & \hspace{0.5cm} +\Bigg. \bigg. \frac{1}{K^{2}(K^{\prime})^{2} \sigma_{h}^{8} \Bar{\beta}^{2} \Bar{\beta}_{c_{1}} \Bar{\beta}_{c_{2}}} \sum_{k_{1},k_{1}^{\prime},k_{2},k_{2}^{\prime}} \!\!\! |h_{IS,c_{1},k^{\prime}_{1}}^{n}(t)|^{2} |h_{IS,c_{2},k^{\prime}_{2}}^{n}(t)|^{2} |h_{c_{1},m_{1},c_{1},k_{1}}^{i_{1},n}(t)|^{2} |h_{c_{2},m_{2},c_{2},k_{2}}^{i_{2},n}(t)|^{2} \bigg) \Bigg], \nonumber
\end{align*}
\begin{align*}
    & = \frac{1}{M^{2}C^{2}} \mathbb{E} \bigg[ \sum_{c_{1}, c_{2}, m_{1}, m_{2}, i_{1}, i_{2}} \Delta \theta_{c_{1},m_{1}}^{i_{1},n}(t) \Delta \theta_{c_{2},m_{2}}^{i_{2},n}(t) \Big( 1 - \frac{\beta_{c_{1},m_{1},c_{1}} \beta_{IS,c_{1}}}{ \Bar{\beta} \Bar{\beta_{c_{1}}} } - \frac{\beta_{c_{2},m_{2},c_{2}} \beta_{IS,c_{2}}}{ \Bar{\beta} \Bar{\beta_{c_{2}}} } \Big. \bigg. \nonumber \\
    & \hspace{0.5cm} \bigg. \Big. + \frac{\beta_{c_{1},m_{1},c_{1}} \beta_{c_{2},m_{2},c_{2}} \beta_{IS,c_{1}} \beta_{IS,c_{2}}}{M C K K^{\prime} I \Bar{\beta}^{2} \Bar{\beta}_{c_{1}} \Bar{\beta}_{c_{2}}} \big( 4\! +\! 2(K^{\prime}\!-\!1) + (M\!-\!1)(K\!-\!1)(I\!-\!1) ( 2 \!+\! (K^{\prime}\!-\!1)(C\!-\!1) ) \big) \Big) \bigg], \nonumber    
\end{align*}
\begin{align}
    & = \mathbb{E} \bigg[ \frac{1}{M^{2}C^{2}} \sum_{c_{1}, c_{2}, m_{1}, m_{2}, i_{1}, i_{2}} \Delta \theta_{c_{1},m_{1}}^{i_{1},n}(t) \Delta \theta_{c_{2},m_{2}}^{i_{2},n}(t) A(m_{1},m_{2},c_{1},c_{2}) \bigg],
\end{align}         
where $A(m_{1},m_{2},c_{1},c_{2})$ is given in Theorem \ref{thm1}. Combining for all symbols, we have
\begin{align}
    & \sum_{n = 1}^{2N} \mathbb{E} \Big[ \big( \Delta \hat{\theta}_{PS,1}^{n}(t)  - \Delta \theta_{PS}^{n}(t) \big)^{2} \Big] \nonumber \\
    & = \frac{1}{M^{2}C^{2}} \sum_{n = 1}^{2N} \sum_{c_{1}, c_{2}, m_{1}, m_{2}, i_{1}, i_{2}} A(m_{1},m_{2},c_{1},c_{2}) \mathbb{E} \Big[ \Delta \theta_{c_{1},m_{1}}^{i_{1},n}(t) \Delta \theta_{c_{2},m_{2}}^{i_{2},n}(t) \Big], \nonumber \\
    & \overset{(a)}{\leq} \frac{\eta^{2}(t) G^{2} I^{2} \tau^{2}}{M^{2}C^{2}} \sum_{c_{1} = 1}^{C} \sum_{c_{2} = 1}^{C} \sum_{m_{1} = 1}^{M} \sum_{m_{2} = 1}^{M} A(m_{1},m_{2},c_{1},c_{2}),
\end{align}
where (a) is obtained using Lemma \ref{intermediary_result1_lemma}.
\end{IEEEproof}
\begin{lemma} \label{lemma12}
    $\mathlarger{\sum_{n = 1}^{2N}} \mathbb{E} \Big[ \big( \Delta \hat{\theta}_{PS,2}^{n}(t) \big)^{2} \Big] \leq \mathlarger{\frac{(K^{\prime}+1) \eta^{2}(t) G^{2} I \tau^{2} }{K K^{\prime}M^{2}C^{2} \Bar{\beta}^{2}} \sum_{c = 1}^{C} \sum_{m = 1}^{M}  \sum_{\substack{m^{\prime} = 1 \\ m^{\prime} \neq m}}^{M} \frac{\beta_{IS,c}^{2} \beta_{c,m,c} \beta_{c,m^{\prime},c}}{\Bar{\beta}_{c}^{2}}}$
\end{lemma}
\begin{IEEEproof}
For $1 \leq n \leq N$, we have
\begin{align}
    & \mathbb{E} \Big[ \big( \Delta \hat{\theta}_{PS,2}^{n}(t) \big)^{2} \Big] = \frac{1}{K^{2}(K^{\prime})^{2}M^{2}C^{2}\sigma_{h}^{8} \Bar{\beta}^{2}} \mathbb{E} \Bigg[ \sum_{c_{1},c_{2},m_{1},m_{2}} \sum_{m_{1}^{\prime} \neq m_{1},m_{2}^{\prime} \neq m_{2}} \sum_{i_{1},i_{2},k_{1},k_{2},k_{1}^{\prime},k_{2}^{\prime}} \Bigg. \nonumber \\
    & \hspace{0.5cm} \Bigg. \times \frac{1}{\Bar{\beta}_{c_{1}} \Bar{\beta}_{c_{2}}}  |h_{PS,c_{1},k_{1}^{\prime}}^{n}(t)|^{2} |h_{PS,c_{2},k_{2}^{\prime}}^{n}(t)|^{2}  \operatorname{Re} \Big\{ (h_{c_{1},m_{1},c_{1},k_{1}}^{i_{1},n}(t))^{\ast} h_{c_{1},m_{1}^{\prime},c_{1},k_{1}}^{i_{1},n}(t) \Delta \theta_{c_{1},m_{1}^{\prime}}^{i_{1},n,cx}(t)  \Big\}  \Bigg. \nonumber \\
    & \hspace{0.5cm} \times \Bigg. \operatorname{Re} \Big\{ (h_{c_{2},m_{2},c_{2},k_{2}}^{i_{2},n}(t))^{\ast} h_{c_{2},m_{2}^{\prime},c_{2},k_{2}}^{i_{2},n}(t) \Delta \theta_{c_{2},m_{2}^{\prime}}^{i_{2},n,cx}(t)  \Big\} \Bigg].
\end{align}
In order for the expectation not to be zero, we need to have $c_{1} \! = \! c_{2}, i_{1} \! = \! i_{2}$ and $k_{1} \! = \! k_{2}$ because of the independence of different channel realizations. Then, using $\mathbb{E} \left[ |h_{IS,c,k}^{n}(t)|^{4} \right] \! = \! 2 \beta_{IS,c}^{2} \sigma_{h}^{4}$, we have
\begin{align*}
    &= \frac{(K^{\prime}+1)}{K^{2}K^{\prime}M^{2}C^{2}\sigma_{h}^{4} \Bar{\beta}^{2}} \mathbb{E} \Bigg[ \sum_{c,m,m^{\prime} \neq m,i,k} \frac{\beta_{IS,c}^{2}}{\Bar{\beta}_{c}^{2}} \bigg( \operatorname{Re} \Big\{ (h_{c,m,c,k}^{i,n}(t))^{\ast} h_{c,m^{\prime},c,k}^{i,n}(t) \Delta \theta_{c,m^{\prime}}^{i,n,cx}(t) \Big\} \bigg)^{2} \Bigg. \nonumber \\
    & \hspace{1cm} + \Bigg. \operatorname{Re} \Big\{ (h_{c,m,c,k}^{i,n}(t))^{\ast} h_{c,m^{\prime},c,k}^{i,n}(t)  \Delta \theta_{c,m^{\prime}}^{i,n,cx}(t) \Big\} \operatorname{Re} \Big\{ (h_{c,m^{\prime},c,k}^{i,n}(t))^{\ast} h_{c,m,c,k}^{i,n}(t) \Delta \theta_{c,m}^{i,n,cx}(t) \Big\} \Bigg], \nonumber     
\end{align*}
\begin{align}
    &= \frac{(K^{\prime}+1)}{2 K K^{\prime}M^{2}C^{2} \Bar{\beta}^{2}} \mathbb{E} \Bigg[ \sum_{c = 1}^{C} \sum_{m = 1}^{M} \sum_{\substack{m^{\prime} = 1 \\ m^{\prime} \neq m}}^{M} \sum_{i = 1}^{I} \frac{\beta_{IS,c}^{2} \beta_{c,m,c} \beta_{c,m^{\prime},c}}{\Bar{\beta}_{c}^{2}} \Bigg. \nonumber \\
    & \hspace{1cm} \times \Bigg. \bigg( \Big( \Delta \theta_{c,m^{\prime}}^{i,n}(t) \Big)^{2} \! + \! \Big( \Delta \theta_{c,m^{\prime}}^{i,n+N}(t) \Big)^{2} \! + \! \Delta \theta_{c,m}^{i,n}(t) \Delta \theta_{c,m^{\prime}}^{i,n}(t) \! - \! \Delta \theta_{c,m}^{i,n+N}(t) \Delta \theta_{c,m^{\prime}}^{i,n+N}(t) \bigg) \Bigg].
\end{align}
For $N+1 \leq n \leq 2N$, we can similarly obtain
\begin{align}
    &= \frac{(K^{\prime}+1)}{2 K K^{\prime}M^{2}C^{2} \Bar{\beta}^{2}} \mathbb{E} \Bigg[  \sum_{c = 1}^{C} \sum_{m = 1}^{M} \sum_{\substack{m^{\prime} = 1 \\ m^{\prime} \neq m}}^{M} \sum_{i = 1}^{I} \frac{\beta_{IS,c}^{2} \beta_{c,m,c} \beta_{c,m^{\prime},c}}{\Bar{\beta}_{c}^{2}}  \Bigg. \nonumber \\ 
    & \hspace{0.5cm} \Bigg. \times \bigg( \Big( \Delta \theta_{c,m^{\prime}}^{i,n}(t) \Big)^{2} \! + \! \Big( \Delta \theta_{c,m^{\prime}}^{i,n-N}(t) \Big)^{2} \! + \! \Delta \theta_{c,m}^{i,n}(t) \Delta \theta_{c,m^{\prime}}^{i,n}(t) \! - \! \Delta \theta_{c,m}^{i,n-N}(t) \Delta \theta_{c,m^{\prime}}^{i,n-N}(t) \bigg) \Bigg].
\end{align}
Combining the two cases, it becomes
\begin{align}
    \sum_{n = 1}^{2N} \mathbb{E} \Big[ \big( \Delta \hat{\theta}_{PS,2}^{n}(t) \big)^{2} \Big] & = \frac{(K^{\prime}+1) }{K K^{\prime}M^{2}C^{2} \Bar{\beta}^{2}} \sum_{c = 1}^{C} \sum_{m = 1}^{M} \sum_{\substack{m^{\prime} = 1 \\ m^{\prime} \neq m}}^{M} \sum_{i = 1}^{I} \frac{\beta_{IS,c}^{2} \beta_{c,m,c} \beta_{c,m^{\prime},c}}{\Bar{\beta}_{c}^{2}} \mathbb{E} \Big[ \big\| \Delta \bm{\theta}_{c,m^{\prime}}^{i}(t) \big\|^{2}_{2} \Big], \nonumber \\
    & \overset{(a)}{\leq} \frac{(K^{\prime}+1) \eta^{2}(t) G^{2} I \tau^{2} }{K K^{\prime}M^{2}C^{2} \Bar{\beta}^{2}} \sum_{c = 1}^{C} \sum_{m = 1}^{M}  \sum_{\substack{m^{\prime} = 1 \\ m^{\prime} \neq m}}^{M} \frac{\beta_{IS,c}^{2} \beta_{c,m,c} \beta_{c,m^{\prime},c}}{\Bar{\beta}_{c}^{2}},
\end{align}
where (a) is obtained using Lemma \ref{intermediary_result2_lemma}.
\end{IEEEproof}
\begin{lemma} \label{lemma13}
    $\mathlarger{\sum_{n = 1}^{2N} \mathbb{E} \Big[ \big( \Delta \hat{\theta}_{PS,3}^{n}(t) \big)^{2} \Big] \leq \frac{(K^{\prime}+1) \eta^{2}(t) G^{2} I \tau^{2} }{K K^{\prime}M^{2}C^{2} \Bar{\beta}^{2}} \sum_{c = 1}^{C} \sum_{\substack{c^{\prime} = 1 \\ c^{\prime} \neq c}}^{C} \sum_{m = 1}^{M} \sum_{m^{\prime} = 1}^{M} \frac{\beta_{IS,c}^{2} \beta_{c,m,c} \beta_{c,m^{\prime},c^{\prime}}}{\Bar{\beta}_{c}^{2}}}$
\end{lemma}
\begin{IEEEproof}
For $1 \leq n \leq N$, we have
\begin{align}
    & \mathbb{E} \Big[ \big( \Delta \hat{\theta}_{PS,3}^{n}(t) \big)^{2} \Big] \nonumber \\
    &= \frac{1}{K^{2}(K^{\prime})^{2}M^{2}C^{2}\sigma_{h}^{8} \Bar{\beta}^{2}} \mathbb{E} \Bigg[ \sum_{c_{1},c_{2}} \sum_{c_{1}^{\prime} \neq c_{1},c_{2}^{\prime} \neq c_{2}} \sum_{m_{1},m_{2}} \sum_{m_{1}^{\prime},m_{2}^{\prime}} \sum_{i_{1},i_{2}} \sum_{k_{1},k_{2},k_{1}^{\prime},k_{2}^{\prime}}  \Bigg. \nonumber \\
    & \hspace{1cm} \Bigg. \times \frac{1}{\Bar{\beta}_{c_{1}} \Bar{\beta}_{c_{2}}} |h_{PS,c_{1},k_{1}^{\prime}}^{n}(t)|^{2} |h_{PS,c_{2},k_{2}^{\prime}}^{n}(t)|^{2} \operatorname{Re} \Big\{ (h_{c_{1},m_{1},c_{1},k_{1}}^{i_{1},n}(t))^{\ast} h_{c_{1},m_{1}^{\prime},c_{1}^{\prime},k_{1}}^{i_{1},n}(t) \Delta \theta_{c_{1}^{\prime},m_{1}^{\prime}}^{i_{1},n,cx}(t) \Big\} \bigg. \Bigg. \nonumber \\
    & \hspace{6cm} \times \Bigg. \bigg. \operatorname{Re} \Big\{ (h_{c_{2},m_{2},c_{2},k_{2}}^{i_{2},n}(t))^{\ast} h_{c_{2},m_{2}^{\prime},c_{2}^{\prime},k_{2}}^{i_{2},n}(t) \big( \Delta \theta_{c_{2}^{\prime},m_{2}^{\prime}}^{i_{2},n,cx}(t) \Big\}  \Bigg].
\end{align}
In order expectation not to be zero, we need to have $c_{1} = c_{2}, c_{1}^{\prime} = c_{2}^{\prime}, m_{1} = m_{2}, m_{1}^{\prime} = m_{2}^{\prime}, i_{1} = i_{2}$ and $k_{1} = k_{2}$ because of the independence of different channel realizations. We get
\begin{align*}
    &= \! \frac{(K^{\prime}+1)}{K^{2}K^{\prime}M^{2}C^{2}\sigma_{h}^{4} \Bar{\beta}^{2}} \mathbb{E} \Bigg[ \! \sum_{c,c^{\prime} \neq c,m,m^{\prime},i,k} \frac{\beta_{IS,c}^{2}}{\Bar{\beta}_{c}^{2}}  \bigg( \! \operatorname{Re} \Big\{ (h_{c,m,c,k}^{i,n}(t))^{\ast} h_{c,m^{\prime},c^{\prime},k}^{i,n}(t) \Delta \theta_{c^{\prime},m^{\prime}}^{i,n,cx}(t) \Big\} \bigg)^{2}  \Bigg], \nonumber     
\end{align*}
\begin{align}
    &= \! \frac{(K^{\prime}+1)}{2 K K^{\prime}M^{2}C^{2} \Bar{\beta}^{2}} \mathbb{E} \! \Bigg[ \! \sum_{c,c^{\prime} \neq c,m,m^{\prime},i} \! \frac{\beta_{IS,c}^{2} \beta_{c,m,c} \beta_{c,m^{\prime},c^{\prime}}}{\Bar{\beta}_{c}^{2}} \! \bigg( \!\! \big( \! \Delta \theta_{c^{\prime},m^{\prime}}^{i,n}(t) \! \big)^{2} \!\! + \!\! \big( \! \Delta \theta_{c^{\prime},m^{\prime}}^{i,n+N}(t) \! \big)^{2} \! \bigg) \! \Bigg].
\end{align}
Similar expression can be obtained for $N+1 \leq n \leq 2N$. Combining two cases, it becomes
\begin{align*}
    & \sum_{n = 1}^{2N} \mathbb{E} \Big[ \big( \Delta \hat{\theta}_{PS,3}^{n}(t) \big)^{2} \Big] \nonumber \\
    &= \! \sum_{n = 1}^{N} \frac{(K^{\prime}+1)}{K K^{\prime}M^{2}C^{2} \Bar{\beta}^{2}} \mathbb{E} \Bigg[ \! \sum_{c,c^{\prime} \neq c,m,m^{\prime},i} \frac{\beta_{IS,c}^{2} \beta_{c,m,c} \beta_{c,m^{\prime},c^{\prime}}}{\Bar{\beta}_{c}^{2}} \bigg( \!\! \big( \! \Delta \theta_{c^{\prime},m^{\prime}}^{i,n}(t) \! \big)^{2} \! + \! \big( \! \Delta \theta_{c^{\prime},m^{\prime}}^{i,n+N}(t) \! \big)^{2} \bigg) \! \Bigg], \nonumber     
\end{align*}
\begin{align}
    & \overset{(a)}{\leq} \frac{(K^{\prime}+1) \eta^{2}(t) G^{2} I \tau^{2} }{K K^{\prime}M^{2}C^{2} \Bar{\beta}^{2}} \sum_{c = 1}^{C} \sum_{\substack{c^{\prime} = 1 \\ c^{\prime} \neq c}}^{C} \sum_{m = 1}^{M} \sum_{m^{\prime} = 1}^{M} \frac{\beta_{IS,c}^{2} \beta_{c,m,c} \beta_{c,m^{\prime},c^{\prime}}}{\Bar{\beta}_{c}^{2}},
\end{align}
where (a) is obtained using Lemma \ref{intermediary_result2_lemma}.
\end{IEEEproof} 
\begin{lemma}
    $ \mathlarger{\sum_{n = 1}^{2N} \mathbb{E} \Big[ \big( \Delta \hat{\theta}_{PS,4}^{n}(t) \big)^{2} \Big] = \frac{(K^{\prime}+1) I \sigma_{z}^{2} N}{KK^{\prime}M^{2}C^{2} P_{t}^{2} \sigma_{h}^{2} \Bar{\beta}^{2}} \sum_{c = 1}^{C} \sum_{m = 1}^{M} \frac{\beta_{IS,c}^{2} \beta_{c,m,c}}{\Bar{\beta}_{c}^{2}} }$
\end{lemma}
\begin{IEEEproof}
For $1 \leq n \leq N$, we have
\begin{align}
    & \mathbb{E} \Big[ \big( \Delta \hat{\theta}_{PS,4}^{n}(t) \big)^{2} \Big] \!\! = \!\! \frac{1}{P_{IS,t}^{2} K^{2} \big(\! K^{\prime} \! \big)^{2} M^{2} C^{2} \sigma_{h}^{8} \Bar{\beta}^{2}} \!  \mathbb{E} \! \bigg[ \! \sum_{c_{1},c_{2}} \sum_{m_{1},m_{2},i_{1},i_{2}} \sum_{k_{1},k_{2},k_{1}^{\prime},k_{2}^{\prime}} \frac{1}{\Bar{\beta}_{c_{1}} \Bar{\beta}_{c_{2}}} \bigg. \nonumber \\
    & \hspace{0.3cm} \bigg. \times | h_{PS,c_{1},k_{1}^{\prime}}^{n}(t)|^{2} |h_{PS,c_{2},k_{2}^{\prime}}^{n}(t)|^{2} \operatorname{Re} \! \Big\{ \!\! \big( h_{c_{1},m_{1},c_{1},k_{1}}^{i_{1},n}(t) \! \big)^{\ast} z_{c_{1},k_{1}}^{i_{1},n}(t) \!\! \Big\} \! \operatorname{Re} \! \Big\{ \!\! \big(h_{c_{2},m_{2},c_{2},k_{2}}^{i_{2},n}(t) \! \big)^{\ast} z_{c_{2},k_{2}}^{i_{2},n}(t) \!\! \Big\} \!\! \bigg].
\end{align}
For a non-zero result, we need to have $c_{1} \! = \! c_{2}, i_{1} \! = \! i_{2}, m_{1} \! = \! m_{2}$ and $k_{1} \! = \! k_{2}$. Then, we get
\begin{align}
    & = \frac{(K^{\prime}+1)}{P_{IS,t}^{2} K^{2}K^{\prime}M^{2}C^{2} \sigma_{h}^{4} \Bar{\beta}^{2}} \mathbb{E} \bigg[ \sum_{c,m,i,k} \frac{\beta_{IS,c}^{2}}{\Bar{\beta}_{c}^{2}} \Big(  \operatorname{Re} \Big\{ (h_{c,m,c,k}^{i,n}(t))^{\ast} z_{c,k}^{i,n}(t) \Big\} \Big)^{2} \bigg], \nonumber \\
    & = \frac{(K^{\prime}+1) I \sigma_{z}^{2}}{2 P_{IS,t}^{2} KK^{\prime}M^{2}C^{2}\sigma_{h}^{2} \Bar{\beta}^{2}} \sum_{c = 1}^{C} \sum_{m = 1}^{M} \frac{\beta_{IS,c}^{2} \beta_{c,m,c}}{\Bar{\beta}_{c}^{2}}.
\end{align}
The derivation is similar for $N+1 \leq n \leq 2N$. Combining the two cases, we get
\begin{equation}
    \sum_{n = 1}^{2N} \mathbb{E} \Big[ \big( \Delta \hat{\theta}_{PS,4}^{n}(t) \big)^{2} \Big] = \frac{(K^{\prime}+1) I \sigma_{z}^{2} N}{P_{IS,t}^{2} KK^{\prime}M^{2}C^{2} \sigma_{h}^{2} \Bar{\beta}^{2}} \sum_{c = 1}^{C} \sum_{m = 1}^{M} \frac{\beta_{IS,c}^{2} \beta_{c,m,c}}{\Bar{\beta}_{c}^{2}}.
\end{equation}
\end{IEEEproof}
\begin{lemma} \label{lemma15} 
    $\mathlarger{\sum_{n = 1}^{2N}} \! \mathbb{E} \! \Big[ \! \big( \! \Delta \hat{\theta}_{PS,5}^{n}(\!t\!) \! \big)^{2} \! \Big] \!\! \leq \!\! \frac{\big( \! 2 \! + \! (M\!-\!1)(C\!-\!2)(K\!-\!1)(I\!-\!1) \!\! \big) \! \eta^{2}(\!t\!) I G^{2} \tau^{2}}{K (K^{\prime}) M^3 C^2 (C-1) \Bar{\beta}^{2}} \mathlarger{\!\! \sum_{c = 1}^{C} \! \sum_{\substack{c^{\prime} = 1 \\ c^{\prime} \neq c}}^{C} \sum_{m_{1} = 1}^{M} \! \sum_{m_{2} = 1}^{M}} \!\! \frac{\beta_{IS,c} \beta_{IS,c^{\prime}} \! \beta_{c^{\prime}\!,m_{1}\!,c^{\prime}} \! \beta_{c^{\prime}\!,m_{2}\!,c^{\prime}}}{\Bar{\beta}_{c^{\prime}}^{2}}$
\end{lemma}
\begin{IEEEproof}
For $1 \leq n \leq N$, the equation becomes
\begin{align}
    & \mathbb{E} \! \Big[ \big( \Delta \hat{\theta}_{PS,5}^{n}(t) \big)^{2} \Big] \!\! = \!\! \frac{1}{K^2 (K^{\prime})^{2} M^2 C^2 \sigma_{h}^{8} \Bar{\beta}^{2}} \mathbb{E} \! \Bigg[  \! \sum_{c_{1},c_{2}} \sum_{c_{1}^{\prime} \neq c_{1}, c_{2}^{\prime} \neq c_{2}} \sum_{m_{1},m_{2},i_{1},i_{2}} \sum_{k_{1},k_{2},k_{1}^{\prime},k_{2}^{\prime}} \Bigg. \nonumber \\
    & \hspace{1cm} \times \Bigg. \frac{1}{\Bar{\beta}_{c_{1}^{\prime}} \Bar{\beta}_{c_{2}^{\prime}}}   |h_{c_{1}^{\prime},m_{1},c_{1}^{\prime},k_{1}}^{i_{1},n}(t)|^{2} |h_{c_{2}^{\prime},m_{2},c_{2}^{\prime},k_{2}}^{i_{2},n}(t)|^{2} \operatorname{Re} \Big\{ (h_{IS,c_{1},k_{1}^{\prime}}^{n}(t))^{\ast} h_{IS,c_{1}^{\prime},k_{1}^{\prime}}^{n}(t) \Delta \theta_{c_{1}^{\prime},m_{1}}^{i_{1},n,cx}(t)  \Big\} \Bigg. \nonumber \\
    & \hspace{7.2cm} \times \Bigg. \operatorname{Re} \Big\{ (h_{IS,c_{2},k_{2}^{\prime}}^{n}(t))^{\ast} h_{IS,c_{2}^{\prime},k_{2}^{\prime}}^{n}(t)  \Delta \theta_{c_{2}^{\prime},m_{2}}^{i_{2},n,cx}(t) \Big\} \Bigg].
\end{align}
For a non-zero answer, we need to have $k_{1}^{\prime} = k_{2}^{\prime}$. The expression becomes
\begin{align}
    = & \frac{\big( 2 + (M-1)(C-2)(K-1)(I-1) \big)}{4 (K^{\prime}) M^3 C^2 (C-1) K I \Bar{\beta}^{2}} \mathbb{E} \Bigg[ \sum_{c,c^{\prime} \neq c,m_{1},m_{2},i_{1},i_{2}} \beta_{IS,c} \beta_{IS,c^{\prime}} \nonumber \\
    & \times \bigg( \frac{\beta_{c^{\prime},m_{1},c^{\prime}} \beta_{c^{\prime},m_{2},c^{\prime}}}{\Bar{\beta}_{c^{\prime}}^{2}} \Big( \Delta \theta_{c^{\prime},m_{1}}^{i_{1},n}(t) \Delta \theta_{c^{\prime},m_{2}}^{i_{2},n}(t) + \Delta \theta_{c^{\prime},m_{1}}^{i_{1},n+N}(t) \Delta \theta_{c^{\prime},m_{2}}^{i_{2},n+N}(t) \Big) \bigg. \Bigg. \nonumber \\
    & \hspace{0.5cm} \Bigg. \bigg. + \frac{\beta_{c^{\prime},m_{1},c^{\prime}} \beta_{c,m_{2},c}}{\Bar{\beta}_{c} \Bar{\beta}_{c^{\prime}}} \Big( \Delta \theta_{c^{\prime},m_{1}}^{i_{1},n}(t) \Delta \theta_{c,m_{2}}^{i_{2},n}(t) - \Delta \theta_{c^{\prime},m_{1}}^{i_{1},n+N}(t) \Delta \theta_{c,m_{2}}^{i_{2},n+N}(t) \Big) \bigg) \Bigg].
\end{align}
The result is similar for $N+1 \leq n \leq 2N$. Overall, it becomes
\begin{align}
    \sum_{n = 1}^{2N} \mathbb{E} \Big[ \big( \Delta \hat{\theta}_{PS,5}^{n}(t) \big)^{2} \Big] & = \frac{\big( 2 + (M-1)(C-2)(K-1)(I-1) \big)}{2 K (K^{\prime}) M^3 C^2 (C-1) I \Bar{\beta}^{2}} \sum_{n = 1}^{N} \sum_{c,c^{\prime} \neq c,m_{1},m_{2},i_{1},i_{2}} \nonumber \\
    & \hspace{-3.5cm} \times \Bigg. \frac{\beta_{IS,c} \beta_{IS,c^{\prime}} \beta_{c^{\prime},m_{1},c^{\prime}} \beta_{c^{\prime},m_{2},c^{\prime}}}{\Bar{\beta}_{c^{\prime}}^{2}} \mathbb{E} \bigg[  \Big( \Delta \theta_{c^{\prime},m_{1}}^{i_{1},n}(t) \Delta \theta_{c^{\prime},m_{2}}^{i_{2},n}(t) + \Delta \theta_{c^{\prime},m_{1}}^{i_{1},n+N}(t) \Delta \theta_{c^{\prime},m_{2}}^{i_{2},n+N}(t) \Big) \bigg], \nonumber \\
    & \hspace{-3.5cm} \overset{(a)}{\leq} \frac{\big( 2 + (M\!-\!1)(C\!-\!2)(K\!-\!1)(I\!-\!1) \big) \eta^{2}(t) I G^{2} \tau^{2}}{K (K^{\prime}) M^3 C^2 (C-1) \Bar{\beta}^{2}} \sum_{c = 1}^{C} \sum_{\substack{c^{\prime} = 1 \\ c^{\prime} \neq c}}^{C} \sum_{m_{1} = 1}^{M} \sum_{m_{2} = 1}^{M} \frac{\beta_{IS,c} \beta_{IS,c^{\prime}} \beta_{c^{\prime},m_{1},c^{\prime}} \beta_{c^{\prime},m_{2},c^{\prime}}}{\Bar{\beta}_{c^{\prime}}^{2}},
\end{align}
where (a) is obtained using Lemma \ref{intermediary_result1_lemma}.
\end{IEEEproof}
\begin{lemma} \label{lemma16} 
    $\mathlarger{\sum_{n = 1}^{2N} \mathbb{E} \Big[ \big( \Delta \hat{\theta}_{PS,6}^{n}(t) \big)^{2} \Big] \leq \frac{\eta^{2}(t) I G^{2} \tau^{2}}{K K^{\prime} M^2 C^2 \Bar{\beta}^{2}} \sum_{c = 1}^{C} \sum_{\substack{c^{\prime} = 1 \\ c^{\prime} \neq c}}^{C} \sum_{m = 1}^{M} \sum_{\substack{m^{\prime} = 1 \\ m^{\prime} \neq m}}^{M} \frac{\beta_{IS,c} \beta_{IS,c^{\prime}} \beta_{c^{\prime},m,c^{\prime}} \beta_{c^{\prime},m^{\prime},c^{\prime}}}{\Bar{\beta}_{c^{\prime}}^{2}}}$
\end{lemma}
\begin{IEEEproof}
For $1 \leq n \leq N$, we have
\begin{align}
    & \mathbb{E} \Big[ \big( \Delta \hat{\theta}_{PS,6}^{n}(t) \big)^{2} \Big]  \nonumber \\
    & = \frac{1}{ K^{2} (K^{\prime})^{2} M^2 C^2 \sigma_{h}^{8} \Bar{\beta}^{2}} \mathbb{E} \Bigg[ \sum_{c_{1},c_{2}} \sum_{c_{1}^{\prime} \neq c_{1},c_{2}^{\prime} \neq c_{2}} \sum_{m_{1},m_{2}} \sum_{m_{1}^{\prime} \neq m_{1},m_{2}^{\prime} \neq m_{2}} \sum_{i_{1},i_{2}} \sum_{k_{1},k_{2},k_{1}^{\prime},k_{2}^{\prime}}  \Bigg. \nonumber \\
    & \hspace{1cm} \times \Bigg. \frac{1}{\Bar{\beta}_{c_{1}^{\prime}} \Bar{\beta}_{c_{2}^{\prime}}} \operatorname{Re} \Big\{ (h_{PS,c_{1},k_{1}^{\prime}}^{n}(t))^{\ast} h_{PS,c_{1}^{\prime},k_{1}^{\prime}}^{n}(t) (h_{c_{1}^{\prime},m_{1},c_{1}^{\prime},k_{1}}^{i_{1},n}(t))^{\ast} h_{c_{1}^{\prime},m_{1}^{\prime},c_{1}^{\prime},k_{1}}^{i_{1},n}(t)  \Delta \theta_{c_{1}^{\prime},m_{1}^{\prime}}^{i_{1},n,cx}(t)  \Big\} \Bigg. \nonumber \\
    & \hspace{2.2cm} \times \Bigg. \operatorname{Re} \Big\{ \! (h_{PS,c_{2},k_{2}^{\prime}}^{n}(t))^{\ast} h_{PS,c_{2}^{\prime},k_{2}^{\prime}}^{n}(t) (h_{c_{2}^{\prime},m_{2},c_{2}^{\prime},k_{2}}^{i_{2},n}(t))^{\ast} h_{c_{2}^{\prime},m_{2}^{\prime},c_{2}^{\prime},k_{2}}^{i_{2},n}(t)  \Delta \theta_{c_{2}^{\prime},m_{2}^{\prime}}^{i_{2},n,cx}(t) \! \Big\} \! \Bigg].
\end{align}
For a non-zero result, we need to have $k_{1} \!\! = \!\! k_{2}$, $k_{1}^{\prime} \!\! = \!\! k_{2}^{\prime}$, $i_{1} \!\! = \!\! i_{2}$, and $c_{1} \!\! = \!\! c_{2}$, which leads to $c_{1}^{\prime} \!\! = \!\! c_{2}^{\prime}$. 
\begin{align}
    & = \frac{1}{ K^{2} (K^{\prime})^{2} M^2 C^2 \sigma_{h}^{8} \Bar{\beta}^{2}} \mathbb{E} \Bigg[ \sum_{c,c^{\prime} \neq c,m,m^{\prime} \neq m,i,k,k^{\prime}} \frac{1}{\Bar{\beta}_{c^{\prime}}^{2}} \Bigg. \nonumber \\
    & \hspace{1cm} \times \Bigg. \Big( \Big( \operatorname{Re} \Big\{ (h_{PS,c,k^{\prime}}^{n}(t))^{\ast} h_{PS,c^{\prime},k^{\prime}}^{n}(t) (h_{c^{\prime},m,c^{\prime},k}^{i,n}(t))^{\ast} h_{c^{\prime},m^{\prime},c^{\prime},k}^{i,n}(t) \Delta \theta_{c^{\prime},m^{\prime}}^{i,n,cx}(t)  \Big\} \Big)^{2} \Big. \Bigg. \nonumber \\
    & \hspace{1.5cm} + \Bigg. \Big. \operatorname{Re} \Big\{ (h_{PS,c,k^{\prime}}^{n}(t))^{\ast} h_{PS,c^{\prime},k^{\prime}}^{n}(t) (h_{c^{\prime},m,c^{\prime},k}^{i,n}(t))^{\ast} h_{c^{\prime},m^{\prime},c^{\prime},k}^{i,n}(t) \Delta \theta_{c^{\prime},m^{\prime}}^{i,n,cx}(t)  \Big\}  \Bigg. \Big. \nonumber \\
    & \hspace{2cm} \times \Bigg. \Big. \operatorname{Re} \Big\{ (h_{PS,c,k^{\prime}}^{n}(t))^{\ast} h_{PS,c^{\prime},k^{\prime}}^{n}(t) (h_{c^{\prime},m^{\prime},c^{\prime},k}^{i,n}(t))^{\ast} h_{c^{\prime},m,c^{\prime},k}^{i,n}(t) \Delta \theta_{c^{\prime},m}^{i,n,cx}(t)   \Big\}  \Big) \Bigg], \nonumber \\
    &  = \mathbb{E} \Bigg[ \sum_{c = 1}^{C} \sum_{\substack{c^{\prime} = 1 \\ c^{\prime} \neq c}}^{C} \sum_{m = 1}^{M} \sum_{\substack{m^{\prime} = 1 \\ m^{\prime} \neq m}}^{M} \sum_{i = 1}^{I} \! \frac{\beta_{IS,c} \beta_{IS,c^{\prime}} \beta_{c^{\prime},m,c^{\prime}} \beta_{c^{\prime},m^{\prime},c^{\prime}}}{ 2 K ( K^{\prime}) M^2 C^2 \Bar{\beta}^{2} \Bar{\beta}_{c^{\prime}}^{2}} \Big(  (\Delta \theta_{c^{\prime},m^{\prime}}^{i,n}(t))^{2} \! + \! (\Delta \theta_{c^{\prime},m^{\prime}}^{i,n+N}(t))^{2}  \Big) \Bigg].
\end{align}
The result is similar for $N+1 \leq n \leq 2N$. Combining the two parts, we have
\begin{align}
    \sum_{n = 1}^{2N} \mathbb{E} \Big[ \big( \Delta \hat{\theta}_{PS,6}^{n}(t) \big)^{2} \Big]
    & = \sum_{c = 1}^{C} \sum_{\substack{c^{\prime} = 1 \\ c^{\prime} \neq c}}^{C} \sum_{m = 1}^{M} \sum_{\substack{m^{\prime} = 1 \\ m^{\prime} \neq m}}^{M} \sum_{i = 1}^{I} \frac{\beta_{IS,c} \beta_{IS,c^{\prime}} \beta_{c^{\prime},m,c^{\prime}} \beta_{c^{\prime},m^{\prime},c^{\prime}}}{ K ( K^{\prime}) M^2 C^2 \Bar{\beta}^{2} \Bar{\beta}_{c^{\prime}}^{2}}  \mathbb{E} \Big[ \big\| \Delta \bm{\theta}_{c^{\prime},m^{\prime}}^{i}(t) \big\|_{2}^{2}  \Big], \nonumber \\
    & \overset{(a)}{\leq} \frac{\eta^{2}(t) I G^{2} \tau^{2}}{K K^{\prime} M^2 C^2 \Bar{\beta}^{2}} \sum_{c = 1}^{C} \sum_{\substack{c^{\prime} = 1 \\ c^{\prime} \neq c}}^{C} \sum_{m = 1}^{M} \sum_{\substack{m^{\prime} = 1 \\ m^{\prime} \neq m}}^{M} \frac{\beta_{IS,c} \beta_{IS,c^{\prime}} \beta_{c^{\prime},m,c^{\prime}} \beta_{c^{\prime},m^{\prime},c^{\prime}}}{\Bar{\beta}_{c^{\prime}}^{2}},
\end{align}
where (a) is obtained using Lemma \ref{intermediary_result2_lemma}.
\end{IEEEproof}
\begin{lemma} \label{lemma17} 
    $ \mathlarger{\sum_{n = 1}^{2N} \mathbb{E} \Big[ \big( \Delta \hat{\theta}_{PS,7}^{n}(t) \big)^{2} \Big] \leq \frac{\eta^{2}(t) I G^{2} \tau^{2}}{K K^{\prime} M^2 C^2 \Bar{\beta}^{2}} \sum_{c = 1}^{C} \sum_{\substack{c^{\prime} = 1 \\ c^{\prime} \neq c}}^{C} \sum_{\substack{c^{\prime \prime} = 1 \\ c^{\prime \prime} \neq c^{\prime}}}^{C} \sum_{m = 1}^{M} \frac{\beta_{IS,c} \beta_{IS,c^{\prime}} \beta_{c^{\prime},m,c^{\prime}} \beta_{c^{\prime},m,c^{\prime \prime}}}{\Bar{\beta}_{c^{\prime}}^{2}}}$
\end{lemma}
\begin{IEEEproof}
For $1 \leq n \leq N$, we have
\begin{align}
    & \mathbb{E} \Big[ \big( \Delta \hat{\theta}_{PS,7}^{n}(t) \big)^{2} \Big] =  \frac{1}{ K^{2} (K^{\prime})^{2} M^2 C^2 \sigma_{h}^{8} \Bar{\beta}^{2}} \nonumber \\
    & \times \mathbb{E} \Bigg[  \sum_{c_{1},c_{2}} \sum_{c_{1}^{\prime} \neq c_{1},c_{2}^{\prime} \neq c_{2}} \sum_{c_{1}^{\prime \prime} \neq c_{1}^{\prime},c_{2}^{\prime \prime} \neq c_{2}^{\prime}} \sum_{m_{1},m_{2}} \sum_{m_{1}^{\prime},m_{2}^{\prime}} \sum_{i_{1},i_{2}} \sum_{k_{1},k_{2},k_{1}^{\prime},k_{2}^{\prime}} \frac{1}{\Bar{\beta}_{c_{1}^{\prime}} \Bar{\beta}_{c_{2}^{\prime}}} \Bigg. \nonumber \\
    & \hspace{0.5cm} \Bigg. \times \operatorname{Re} \Big\{ (h_{PS,c_{1},k_{1}^{\prime}}^{n}(t))^{\ast} h_{PS,c_{1}^{\prime},k_{1}^{\prime}}^{n}(t) (h_{c_{1}^{\prime},m_{1},c_{1}^{\prime},k_{1}}^{i_{1},n}(t))^{\ast} h_{c_{1}^{\prime},m_{1}^{\prime},c_{1}^{\prime \prime},k_{1}}^{i_{1},n}(t) \Delta \theta_{c_{1}^{\prime \prime},m_{1}^{\prime}}^{i_{1},n,cx}(t) \Big\} \Bigg. \nonumber \\
    & \hspace{0.5cm} \Bigg. \times \operatorname{Re} \Big\{ (h_{PS,c_{2},k_{2}^{\prime}}^{n}(t))^{\ast} h_{PS,c_{2}^{\prime},k_{2}^{\prime}}^{n}(t) (h_{c_{2}^{\prime},m_{2},c_{2}^{\prime},k_{2}}^{i_{2},n}(t))^{\ast} h_{c_{2}^{\prime},m_{2}^{\prime},c_{2}^{\prime \prime},k_{2}}^{i_{2},n}(t) \Delta \theta_{c_{2}^{\prime \prime},m_{2}^{\prime}}^{i_{2},n,cx}(t) \Big\} \Bigg].
\end{align}
For a non-zero answer, we need to have $m_{1} = m_{2}$, $m_{1}^{\prime} = m_{2}^{\prime} $ $k_{1} = k_{2}$, $k_{1}^{\prime} = k_{2}^{\prime}$, $i_{1} = i_{2}$, and $c_{1}^{\prime} = c_{2}^{\prime}$ and $c_{1}^{\prime \prime} = c_{2}^{\prime \prime}$, which leads to $c_{1} = c_{2}$. Then, we have
\begin{align}
    & = \frac{1}{ K^{2} (K^{\prime})^{2} M^2 C^2 \sigma_{h}^{8} \Bar{\beta}^{2}} \mathbb{E} \Bigg[ \sum_{c,c^{\prime} \neq c,c^{\prime \prime} \neq c^{\prime}} \sum_{m,m^{\prime},i,k,k^{\prime}} \frac{1}{\Bar{\beta}_{c^{\prime}}^{2}} \Bigg. \nonumber \\
    & \hspace{1cm} \times \Bigg. \Big( \operatorname{Re} \Big\{ (h_{PS,c,k^{\prime}}^{n}(t))^{\ast} h_{PS,c^{\prime},k^{\prime}}^{n}(t) (h_{c^{\prime},m,c^{\prime},k}^{i,n}(t))^{\ast} h_{c^{\prime},m^{\prime},c^{\prime \prime},k}^{i,n}(t) \Delta \theta_{c^{\prime \prime},m}^{i,n,cx}(t) \Big\} \Big)^{2}  \Bigg], \nonumber \\
    & = \! \mathbb{E} \! \Bigg[ \sum_{c,c^{\prime} \neq c,c^{\prime \prime} \neq c^{\prime}} \sum_{m,m^{\prime},i} \! \frac{\beta_{IS,c} \beta_{IS,c^{\prime}} \beta_{c^{\prime}\!,m,c^{\prime}} \beta_{c^{\prime}\!,m^{\prime},c^{\prime \prime}\!}}{2 K K^{\prime} M^2 C^2 \Bar{\beta}^{2} \Bar{\beta}_{c^{\prime}}^{2}} \! \Big( \! (\Delta \theta_{c^{\prime \prime},m^{\prime}}^{i,n}(t))^{2} + (\Delta \theta_{c^{\prime \prime},m^{\prime}}^{i,n+N}(t))^{2}  \Big) \! \Bigg].
\end{align}
The derivation is similar for $N+1 \leq n \leq 2N$. Combining two parts, we have
\begin{align}
    & \sum_{n = 1}^{2N} \mathbb{E} \Big[ \big( \Delta \hat{\theta}_{PS,7}^{n}(t) \big)^{2} \Big] \nonumber \\
    & = \sum_{n = 1}^{N} \sum_{c,c^{\prime} \neq c,c^{\prime \prime} \neq c^{\prime}} \sum_{m,m^{\prime},i} \frac{\beta_{IS,c} \beta_{IS,c^{\prime}} \beta_{c^{\prime},m,c^{\prime}} \beta_{c^{\prime},m^{\prime},c^{\prime \prime}}}{K ( K^{\prime}) M^2 C^2 \Bar{\beta}^{2} \Bar{\beta}_{c^{\prime}}^{2}} \mathbb{E} \Big[ (\Delta \theta_{c^{\prime \prime},m^{\prime}}^{i,n}(t))^{2} + (\Delta \theta_{c^{\prime \prime},m^{\prime}}^{i,n+N}(t))^{2} \Big], \nonumber \\
    & \overset{(a)}{\leq} \frac{\eta^{2}(t) I G^{2} \tau^{2}}{K K^{\prime} M^2 C^2 \Bar{\beta}^{2}} \sum_{c = 1}^{C} \sum_{\substack{c^{\prime} = 1 \\ c^{\prime} \neq c}}^{C} \sum_{\substack{c^{\prime \prime} = 1 \\ c^{\prime \prime} \neq c^{\prime}}}^{C} \sum_{m = 1}^{M} \frac{\beta_{IS,c} \beta_{IS,c^{\prime}} \beta_{c^{\prime},m,c^{\prime}} \beta_{c^{\prime},m,c^{\prime \prime}}}{\Bar{\beta}_{c^{\prime}}^{2}}.
\end{align}
where (a) is due to Lemma \ref{intermediary_result2_lemma}.
\end{IEEEproof}
\begin{lemma} \label{lemma18} 
    $\mathlarger{\sum_{n = 1}^{2N} \mathbb{E} \Big[ \Big( \Delta \hat{\theta}_{PS,6}^{n}(t) \Big)^{2} \Big] = \frac{\sigma_{z}^{2} I N}{P_{IS,t}^{2} K (K^{\prime}) M^{2} C^{2} \sigma_{h}^{2} \Bar{\beta}^{2}} \sum_{c = 1}^{C} \sum_{\substack{c^{\prime} = 1 \\ c^{\prime} \neq c}}^{C}  \sum_{m = 1}^{M} \frac{ \beta_{IS,c} \beta_{IS,c^{\prime}} \beta_{c^{\prime},m,c^{\prime}}}{\Bar{\beta}_{c^{\prime}}^{2}}}$
\end{lemma}
\begin{IEEEproof}
For $1 \leq n \leq N$, we have
\begin{align}
    & \mathbb{E} \Big[ \Big( \Delta \hat{\theta}_{PS,6}^{n}(t) \Big)^{2} \Big] = \frac{1}{P_{IS,t}^{2} K^2 (K^{\prime})^{2} M^2 C^2 \sigma_{h}^{8} \Bar{\beta}^{2}} \mathbb{E} \Bigg[ \sum_{c_{1},c_{2}} \sum_{c_{1}^{\prime} \neq c_{1},c_{2}^{\prime} \neq c_{2}} \sum_{m_{1},m_{2},i_{1},i_{2}} \sum_{k_{1},k_{2},k_{1}^{\prime},k_{2}^{\prime}} \nonumber \\
    & \hspace{3.1cm} \times \frac{1}{\Bar{\beta}_{c_{1}^{\prime}} \Bar{\beta}_{c_{2}^{\prime}}} \operatorname{Re} \Big\{ \big( h_{PS,c_{1},k_{1}^{\prime}}^{n}(t) \big)^{\ast} h_{PS,c_{1}^{\prime},k_{1}^{\prime}}^{n}(t) \big( h_{c_{1}^{\prime},m_{1},c_{1}^{\prime},k_{1}}^{i_{1},n}(t) \big)^{\ast} z_{IS,c_{1}^{\prime},k_{1}}^{i_{1},n}(t)  \Big\} \Bigg. \nonumber \\
    & \hspace{4.2cm} \times \Bigg. \operatorname{Re} \Big\{ \big( h_{PS,c_{2},k_{2}^{\prime}}^{n}(t) \big)^{\ast} h_{PS,c_{2}^{\prime},k_{2}^{\prime}}^{n}(t) \big( h_{c_{2}^{\prime},m_{2},c_{2}^{\prime},k_{2}}^{i_{2},n}(t) \big)^{\ast} z_{IS,c_{2}^{\prime},k_{2}}^{i_{2},n}(t)  \Big\} \Bigg].
\end{align}
For a non-zero answer, we have $m_{1} \!\! = \!\! m_{2}, c_{1} \!\! = \!\! c_{2}, c_{1}^{\prime} \!\! = \!\! c_{2}^{\prime}, k_{1} \!\! = \!\! k_{2}, k_{1}^{\prime} \!\! = \!\! k_{2}^{\prime}, i_{1} \!\! = \!\! i_{2}$. Then, it becomes
\begin{align}
    \mathbb{E} \Big[ \Big( \Delta \hat{\theta}_{PS,6}^{n}(t) \Big)^{2} \Big] &= \frac{\sigma_{z}^{2} I }{2 P_{IS,t}^{2} K (K^{\prime}) M^{2} C^{2} \sigma_{h}^{2} \Bar{\beta}^{2}} \sum_{c = 1}^{C} \sum_{\substack{c^{\prime} = 1 \\ c^{\prime} \neq c}}^{C} \sum_{m = 1}^{M} \frac{\big( \beta_{IS,c} \beta_{IS,c^{\prime}} \beta_{c^{\prime},m,c^{\prime}} \big)}{\Bar{\beta}_{c^{\prime}}^{2}}.
\end{align}
The solution is the same for $N \! + \! 1 \! \leq \! n \! \leq \! 2N$. Adding all the terms concludes the lemma.
\end{IEEEproof}
\begin{lemma} \label{lemma19}
    $\mathlarger{\sum_{n = 1}^{2N} \mathbb{E} \Big[ \Big( \Delta \hat{\theta}_{PS,7}^{n}(t) \Big)^{2} \Big] = \frac{\sigma_{z}^{2} N }{ P_{IS,t}^{2} (K^{\prime}) C^{2} \sigma_{h}^{2} \Bar{\beta}^{2}} \sum_{c = 1}^{C} \beta_{IS,c}}$
\end{lemma}
\begin{IEEEproof}
For $1 \leq n \leq N$, we have
\begin{align}
    \mathbb{E} \Big[ \Big( \Delta \hat{\theta}_{PS,7}^{n}(t) \Big)^{2} \Big] = \frac{1}{P_{IS,t}^{2} (K^{\prime})^{2} C^2 \sigma_{h}^{4} \Bar{\beta}^{2}} \mathbb{E} &\Bigg[ \sum_{c_{1} = 1}^{C} \sum_{c_{2} = 1}^{C}  \sum_{k_{1}^{\prime} = 1}^{K^{\prime}} \sum_{k_{2}^{\prime} = 1}^{K^{\prime}}  \operatorname{Re} \Big\{ \big( h_{PS,c_{1},k_{1}^{\prime}}^{n}(t) \big)^{\ast}  z_{PS,k_{1}^{\prime}}^{n}(t)  \Big\} \nonumber \\
    & \hspace{0.5cm} \times \operatorname{Re} \Big\{ \big( h_{PS,c_{2},k_{2}^{\prime}}^{n}(t) \big)^{\ast}  z_{PS,k_{2}^{\prime}}^{n}(t)  \Big\} \Bigg].
\end{align}
For a non-zero answer, we have $c_{1} \! = \! c_{2}$ and $k_{1}^{\prime} \! = \! k_{2}^{\prime}$. Then, it becomes
\begin{align}
    \mathbb{E} \Big[ \Big( \Delta \hat{\theta}_{PS,7}^{n}(t) \Big)^{2} \Big] & = \frac{1}{P_{IS,t}^{2} (K^{\prime})^{2} C^2 \sigma_{h}^{4} \Bar{\beta}^{2}} \mathbb{E} \Bigg[ \sum_{c = 1}^{C} \sum_{k^{\prime} = 1}^{K^{\prime}} \Big( \operatorname{Re} \Big\{ (h_{PS,c,k^{\prime}}^{n}(t))^{\ast}  z_{PS,k^{\prime}}^{n}(t)  \Big\} \Big)^{2}  \Bigg], \nonumber \\
    & = \frac{\sigma_{z}^{2}}{2 P_{IS,t}^{2} (K^{\prime}) C^{2} \sigma_{h}^{2} \Bar{\beta}^{2}} \sum_{c = 1}^{C} \beta_{IS,c}.
\end{align}
The solution is similar for $N \! + \! 1 \! \leq \! n \! \leq \! 2N$. Summing over all the symbols concludes the lemma.
\end{IEEEproof}
Combining Lemmas \ref{lemma11}-\ref{lemma19} completes the proof of Lemma \ref{convproof1}.
\section*{Appendix C: Proof of Lemma \ref{convproof2}}
We have 
\begin{align}
    \mathbb{E} \! \big[  \big\|  v(t+1) \! - \! \bm{\theta}^{*}  \big\|_{2}^{2}  \big] \!\! &= \!\! \mathbb{E} \! \big[  \big\|  \bm{\theta}_{PS}(t) \! + \! \Delta \bm{\theta}_{PS}(t) \! - \! \bm{\theta}^{*} \big\|_{2}^{2}  \big], \nonumber \\
    & = \! \mathbb{E} \! \big[ \big\| \bm{\theta}_{PS}(t) \! - \! \bm{\theta}^{*} \big\|_{2}^{2} \big] \! + \! \mathbb{E} \! \big[ \big\|  \Delta \bm{\theta}_{PS}(t) \big\|_{2}^{2} \big] \! + \! 2 \mathbb{E} \! \big[ \langle \bm{\theta}_{PS}(t) \! - \! \bm{\theta}^{*}, \Delta \bm{\theta}_{PS}(t) \rangle \big]. \label{eq: lemma2initial}
\end{align}
Where the second term can be bounded as
\begin{align}
    & \hspace{-2.7cm} \mathbb{E} \big[ \big\|  \Delta \bm{\theta}_{PS}(t) \big\|_{2}^{2} \big] = \mathbb{E} \bigg[ \Big\| \frac{1}{MC} \sum_{c = 1}^{C} \sum_{m = 1}^{M} \sum_{i = 1}^{I} \Delta \bm{\theta}_{c,m}^{i}(t) \Big\|_{2}^{2} \bigg] \nonumber \\
    &  \overset{(a)}{\leq} \frac{1}{MC} \sum_{c = 1}^{C} \sum_{m = 1}^{M} \sum_{i = 1}^{I} \mathbb{E} \big[ \big\| \Delta \bm{\theta}_{c,m}^{i}(t) \big\|_{2}^{2} \big] \nonumber \\
    & \overset{(b)}{=} \frac{\eta^{2}(t)}{MC} \sum_{c = 1}^{C} \sum_{m = 1}^{M} \sum_{i = 1}^{I}  \mathbb{E} \bigg[ \Big\| \sum_{j = 1}^{\tau} \nabla F_{c,m} \big( \bm{\theta}_{c,m}^{i,j}(t), \bm{\xi}_{c,m}^{i,j}(t) \big) \Big\|_{2}^{2} \bigg] \nonumber \\
    &  \leq \frac{\eta^{2}(t) \tau}{MC} \sum_{c = 1}^{C} \sum_{m = 1}^{M} \sum_{i = 1}^{I} \sum_{j = 1}^{\tau} \mathbb{E} \big[ \big\| \nabla F_{c,m} \big( \bm{\theta}_{c,m}^{i,j}(t), \bm{\xi}_{c,m}^{i,j}(t) \big) \big\|_{2}^{2} \big] \nonumber \\
    &  \overset{(c)}{\leq} \eta^{2}(t) I \tau^{2} G^{2},
\end{align}
where (a) is due to the convexity of $\left\|  \right\|_{2}^{2}$, (b) comes from utilizing (\ref{eq: useriteration}), and (c) is obtained using Assumption \ref{assumption2}. Plugging in the result to (\ref{eq: lemma2initial}), we have
\begin{equation}
    \mathbb{E} \big[ \big\| v(t+1) - \bm{\theta}^{*} \big\|_{2}^{2} \big] \leq \mathbb{E} \big[ \big\| \bm{\theta}_{PS}(t) - \bm{\theta}^{*} \big\|_{2}^{2} \big] + \eta^{2}(t) I \tau^{2} G^{2} + 2 \mathbb{E} \big[ \langle \bm{\theta}_{PS}(t) - \bm{\theta}^{*}, \Delta \bm{\theta}_{PS}(t) \rangle \big]. \label{eq: lemma2initial2}
\end{equation}
The last term of \eqref{eq: lemma2initial2}, we have
\begin{align}
    & 2 \mathbb{E} \big[ \langle \bm{\theta}_{PS}(t) - \bm{\theta}^{*}, \Delta \bm{\theta}_{PS}(t) \rangle \big] = \frac{2}{MC} \sum_{c = 1}^{C} \sum_{m = 1}^{M} \sum_{i = 1}^{I} \mathbb{E} \big[ \langle \bm{\theta}_{PS}(t) - \bm{\theta}^{*}, \Delta \bm{\theta}_{c,m}^{i}(t) \rangle \big] \nonumber \\
    & \hspace{1.5cm} = \frac{2 \eta(t)}{MC} \sum_{c = 1}^{C} \sum_{m = 1}^{M} \sum_{i = 1}^{I} \mathbb{E} \Big[ \langle \bm{\theta}^{*} - \bm{\theta}_{PS}(t) , \sum_{j = 1}^{\tau} \nabla F_{c,m} \big( \bm{\theta}_{c,m}^{i,j}(t), \bm{\xi}_{c,m}^{i,j}(t) \big) \rangle \Big] \nonumber \\
    & \hspace{1.5cm} = \frac{2 \eta(t)}{MC} \sum_{c = 1}^{C} \sum_{m = 1}^{M} \sum_{i = 1}^{I} \mathbb{E} \big[ \langle \bm{\theta}^{*} - \bm{\theta}_{PS}(t), \nabla F_{c,m} \big( \bm{\theta}_{PS}(t), \bm{\xi}_{c,m}^{1,1}(t) \big) \rangle \big] \nonumber \\
    & \hspace{2cm} + \frac{2 \eta(t)}{MC} \sum_{c = 1}^{C} \sum_{m = 1}^{M} \sum_{i = 1}^{I} \mathbb{E} \Big[ \langle \bm{\theta}^{*} - \bm{\theta}_{PS}(t), \sum_{j = 2}^{\tau} \nabla F_{c,m} \big( \bm{\theta}_{c,m}^{i,j}(t), \bm{\xi}_{c,m}^{i,j}(t) \big) \rangle \Big]. \label{eq: lemma2initial3}
\end{align}
For the first term of \eqref{eq: lemma2initial3}, we have
\begin{align}
    & \frac{2 \eta(t)}{MC} \sum_{c = 1}^{C} \sum_{m = 1}^{M} \sum_{i = 1}^{I} \mathbb{E} \big[ \langle \bm{\theta}^{*} - \bm{\theta}_{PS}(t), \nabla F_{c,m} \big( \bm{\theta}_{PS}(t), \bm{\xi}_{c,m}^{1,1}(t) \big) \rangle \big] \nonumber \\
    & \hspace{2cm} \overset{(a)}{=} \frac{2 \eta(t)}{MC} \sum_{c = 1}^{C} \sum_{m = 1}^{M} \sum_{i = 1}^{I} \mathbb{E} \big[ \langle \bm{\theta}^{*} - \bm{\theta}_{PS}(t), \nabla F_{c,m} \big( \bm{\theta}_{PS}(t) \big) \rangle \big] \nonumber \\
    & \hspace{2cm} \overset{(b)}{\leq} \frac{2 \eta(t)}{MC} \sum_{c = 1}^{C} \sum_{m = 1}^{M} \sum_{i = 1}^{I} \mathbb{E} \big[ F_{c,m}(\bm{\theta}^{*}) - F_{c,m}(\bm{\theta}_{PS}(t)) - \frac{\mu}{2} \big\| \bm{\theta}_{PS}(t) - \bm{\theta}^{*} \big\|_{2}^{2} \big] \nonumber \\
    & \hspace{2cm} = 2 \eta(t) I \Big( F^{*} - \mathbb{E} \big[ F(\bm{\theta}_{PS}(t)) \big] - \frac{\mu}{2} \mathbb{E} \big[ \big\| \bm{\theta}_{PS}(t) - \bm{\theta}^{*} \big\|_{2}^{2} \big] \Big) \nonumber \\
    & \hspace{2cm} \overset{(c)}{\leq} - \eta(t) I \mu \mathbb{E} \big[ \big\| \bm{\theta}_{PS}(t) - \bm{\theta}^{*} \big\|_{2}^{2} \big], \label{eq: lemma2initial4}
\end{align}
where (a) comes from $\mathbb{E}_{\xi} \left[ \! \nabla F_{c,m} \left( \! \bm{\theta}_{PS}(t) , \bm{\xi}_{c,m}^{1,1}(t) \! \right) \! \right] \!\! = \!\! \nabla F_{c,m}(\bm{\theta}_{PS}(t))$, (b) holds due to the $\mu$-strong convexity of $F_{c,m}$, and (c) follows since $F^{*} \! \leq \! F(\bm{\theta}(t))$. For the second term of (\ref{eq: lemma2initial3}), we have
\begin{align*}
    & \frac{2 \eta(t)}{MC} \sum_{c = 1}^{C} \sum_{m = 1}^{M} \sum_{i = 1}^{I} \mathbb{E} \Big[ \langle \bm{\theta}^{*} - \bm{\theta}_{PS}(t), \sum_{j = 2}^{\tau} \nabla F_{c,m} \big( \bm{\theta}_{c,m}^{i,j}(t), \bm{\xi}_{c,m}^{i,j}(t) \big) \rangle \Big] \nonumber \\
    \hspace{1.5cm} & = \frac{2 \eta(t)}{MC} \sum_{c = 1}^{C} \sum_{m = 1}^{M} \sum_{i = 1}^{I} \sum_{j = 2}^{\tau} \mathbb{E} \big[ \langle \bm{\theta}^{*} - \bm{\theta}_{PS}(t), \nabla F_{c,m} \big( \bm{\theta}_{c,m}^{i,j}(t), \bm{\xi}_{c,m}^{i,j}(t) \big) \rangle \big] \nonumber     
\end{align*}
\begin{align}
    \hspace{1.5cm} & = \frac{2 \eta(t)}{MC} \sum_{c = 1}^{C} \sum_{m = 1}^{M} \sum_{i = 1}^{I} \sum_{j = 2}^{\tau} \mathbb{E} \big[ \langle \bm{\theta}_{c,m}^{i,j}(t) - \bm{\theta}_{PS}(t), \nabla F_{c,m} \big( \bm{\theta}_{c,m}^{i,j}(t), \bm{\xi}_{c,m}^{i,j}(t) \big) \rangle \big] \nonumber \\
    \hspace{2cm} & + \frac{2 \eta(t)}{MC} \sum_{c = 1}^{C} \sum_{m = 1}^{M} \sum_{i = 1}^{I} \sum_{j = 2}^{\tau} \mathbb{E} \big[ \langle \bm{\theta}^{*} - \bm{\theta}_{c,m}^{i,j}(t), \nabla F_{c,m} \big( \bm{\theta}_{c,m}^{i,j}(t), \bm{\xi}_{c,m}^{i,j}(t) \big) \rangle \big]. \label{eq: lemma2initial5}
\end{align}
Using Cauchy-Schwarz inequality, we obtain
\begin{align}
    & \frac{2 \eta(t)}{MC} \sum_{c = 1}^{C} \sum_{m = 1}^{M} \sum_{i = 1}^{I} \sum_{j = 2}^{\tau} \mathbb{E} \big[ \langle \bm{\theta}_{c,m}^{i,j}(t) - \bm{\theta}_{PS}(t), \nabla F_{c,m} \big( \bm{\theta}_{c,m}^{i,j}(t), \bm{\xi}_{c,m}^{i,j}(t) \big) \rangle \big] \nonumber \\
    & \hspace{1cm} \leq \frac{\eta(t)}{MC} \sum_{c = 1}^{C} \sum_{m = 1}^{M} \sum_{i = 1}^{I} \sum_{j = 2}^{\tau} \mathbb{E} \bigg[ \frac{1}{\eta(t)} \big\| \bm{\theta}_{c,m}^{i,j}(t) - \bm{\theta}_{PS}(t) \big\|_{2}^{2} + \eta(t) \big\| \nabla F_{c,m} \big( \bm{\theta}_{c,m}^{i,j}(t), \bm{\xi}_{c,m}^{i,j}(t) \big) \big\|_{2}^{2} \bigg] \nonumber \\
    & \hspace{1cm} \overset{(a)}{\leq} \frac{1}{MC} \sum_{c = 1}^{C} \sum_{m = 1}^{M} \sum_{i = 1}^{I} \sum_{j = 2}^{\tau} \mathbb{E} \bigg[ \frac{1}{\eta(t)} \big\| \bm{\theta}_{c,m}^{i,j}(t) - \bm{\theta}_{PS}(t) \big\|_{2}^{2} \bigg] + \eta^{2}(t) I (\tau - 1) G^{2}, \label{eq: lemma2initial6}
\end{align}
where (a) is obtained using Assumption \ref{assumption2}. The following lemma will give an upper bound on the second term of (\ref{eq: lemma2initial5}).
\begin{lemma} \label{lemma2aproof}
$\mathlarger{\frac{2 \eta(t)}{MC} \sum_{c = 1}^{C} \sum_{m = 1}^{M} \sum_{i = 1}^{I} \sum_{j = 2}^{\tau}} \mathbb{E} \big[ \langle \bm{\theta}^{*} - \bm{\theta}_{c,m}^{i,j}(t), \nabla F_{c,m} \big( \bm{\theta}_{c,m}^{i,j}(t), \bm{\xi}_{c,m}^{i,j}(t) \big) \rangle \big] \\
    ~~~~~~~~~~~~~~~~ = - \mu \eta(t) (1 - \eta(t)) I (\tau - 1) \mathbb{E} \big[ \big\| \bm{\theta}_{PS}(t) - \bm{\theta}^{*} \big\|_{2}^{2} \big] \\
    ~~~~~~~~~~~~~~~~~~~~ + \mathlarger{\frac{\mu (1 - \eta(t))}{MC} \sum_{c = 1}^{C} \sum_{m = 1}^{M} \sum_{i = 1}^{I} \sum_{j = 2}^{\tau}} \mathbb{E} \big[ \big\| \bm{\theta}_{c,m}^{i,j}(t) - \bm{\theta}_{PS}(t) \big\|_{2}^{2} \big] + 2 \eta(t) I (\tau - 1) \Gamma.$
\end{lemma}
\begin{IEEEproof}
See Appendix D.
\end{IEEEproof}
Using the results in (\ref{eq: lemma2initial6}) and (\ref{eq: lemma2aproof2}), we can write (\ref{eq: lemma2initial5}) as
\begin{align}
    & \frac{2 \eta(t)}{MC} \sum_{c = 1}^{C} \sum_{m = 1}^{M} \sum_{i = 1}^{I} \mathbb{E} \Big[ \langle \bm{\theta}^{*} - \bm{\theta}_{PS}(t), \sum_{j = 2}^{\tau} \nabla F_{c,m} \big( \bm{\theta}_{c,m}^{i,j}(t), \bm{\xi}_{c,m}^{i,j}(t) \big) \rangle \Big] \nonumber \\
    & \hspace{1cm} = - \mu \eta(t) (1 - \eta(t)) I (\tau - 1) \mathbb{E} \big[ \big\| \bm{\theta}_{PS}(t) - \bm{\theta}^{*} \big\|_{2}^{2} \big] \nonumber \\
    & \hspace{1.5cm} + \frac{( 1 + \mu (1 - \eta(t)))}{MC} \sum_{c = 1}^{C} \sum_{m = 1}^{M} \sum_{i = 1}^{I} \sum_{j = 2}^{\tau} \mathbb{E} \big[ \big\| \bm{\theta}_{c,m}^{i,j}(t) - \bm{\theta}_{PS}(t) \big\|_{2}^{2} \big] \nonumber \\
    & \hspace{1.5cm} + \eta^{2}(t) I (\tau - 1) G^{2} + 2 \eta(t) I (\tau - 1) \Gamma. \label{eq: lemma2initial7}
\end{align}
Also, we have
\begin{align}
    \frac{1}{MC} \! \sum_{c = 1}^{C} \sum_{m = 1}^{M} \sum_{i = 1}^{I} \sum_{j = 2}^{\tau} \! \mathbb{E} \! \big[  \big\| \bm{\theta}_{c,m}^{i,j}(t) \!\! - \!\! \bm{\theta}_{PS}(t) \big\|_{2}^{2}  \big] \!\! & = \!\! \frac{\eta^{2}}{MC} \!  \sum_{c = 1}^{C} \sum_{m = 1}^{M} \sum_{i = 1}^{I} \sum_{j = 2}^{\tau} \! \mathbb{E} \! \big[ \big\| \nabla F_{c,m} \! \big( \! \bm{\theta}_{c,m}^{i,j}(t), \bm{\xi}_{c,m}^{i,j}(t) \! \big) \big\|_{2}^{2} \big] \nonumber \\
    & \overset{(a)}{\leq} \eta^{2} I G^{2} \frac{\tau (\tau - 1) (2 \tau - 1)}{6},
\end{align}
where (a) is due to the convexity of $L_{2}$ norm and Assumption \ref{assumption2}. For $\eta(t) \leq 1$, we have
\begin{align}
    & \frac{2 \eta(t)}{MC} \sum_{c = 1}^{C} \sum_{m = 1}^{M} \sum_{i = 1}^{I} \mathbb{E} \bigg[ \langle \bm{\theta}^{*} - \bm{\theta}_{PS}(t), \sum_{j = 2}^{\tau} \nabla F_{c,m} \big( \bm{\theta}_{c,m}^{i,j}(t), \bm{\xi}_{c,m}^{i,j}(t) \big) \rangle \bigg] \nonumber \\
    & \hspace{0.75cm} \leq \mu \eta(t) (1 \! - \! \eta(t)) I (\tau \! - \! 1) \mathbb{E} \big[ \big\| \bm{\theta}_{PS}(t) \! - \! \bm{\theta}^{*} \big\|_{2}^{2} \big] \! + \! \big( 1 \! + \! \mu (1 \! - \! \eta(t)) \big) \eta^{2}(t) I G^{2} \frac{\tau (\tau \! - \! 1) (2 \tau \! - \! 1)}{6} \nonumber \\
    & \hspace{1.25cm}  \! + \! \eta^{2}(t) I (\tau \! - \! 1) G^{2} \! + \! 2 \eta(t) I (\tau \! - \! 1) \Gamma. \label{eq: lemma2initial8}
\end{align}
Substituting the results in (\ref{eq: lemma2initial4}) and (\ref{eq: lemma2initial8}) into (\ref{eq: lemma2initial3}), we get
\begin{align}
    2 \mathbb{E} & \big[ \langle \bm{\theta}_{PS}(t) - \bm{\theta}^{*}, \Delta \bm{\theta}_{PS}(t) \rangle \big] \leq \mu \eta(t) I (\tau - \eta(t)(\tau - 1)) \mathbb{E} \big[ \big\| \bm{\theta}_{PS}(t) - \bm{\theta}^{*} \big\|_{2}^{2} \big] \nonumber \\
    & \hspace{0.5cm} \! + \! \big( 1 \! + \! \mu (1 \! - \! \eta(t)) \big) \eta^{2}(t) I G^{2} \frac{\tau (\tau \! - \! 1) (2 \tau \! - \! 1)}{6} + \eta^{2}(t) I (\tau \! - \! 1) G^{2} \! + \! 2 \eta(t) I (\tau \! - \! 1) \Gamma. \label{eq: lemma2final}
\end{align}
Lemma \ref{convproof2} is concluded by plugging (\ref{eq: lemma2final}) into (\ref{eq: lemma2initial2}).
\section*{Appendix D: Proof of Lemma \ref{lemma2aproof}}
We have
\begin{align}
    & \frac{2 \eta(t)}{MC} \sum_{c = 1}^{C} \sum_{m = 1}^{M} \sum_{i = 1}^{I} \sum_{j = 2}^{\tau} \mathbb{E} \big[ \langle \bm{\theta}^{*} - \bm{\theta}_{c,m}^{i,j}(t), \nabla F_{c,m} \big( \bm{\theta}_{c,m}^{i,j}(t), \bm{\xi}_{c,m}^{i,j}(t) \big) \rangle \big] \nonumber \\
    & \hspace{0.3cm} \overset{(a)}{\leq} \frac{2 \eta(t)}{MC} \sum_{c = 1}^{C} \sum_{m = 1}^{M} \sum_{i = 1}^{I} \sum_{j = 2}^{\tau} \mathbb{E}  \big[ \langle \bm{\theta}^{*} - \bm{\theta}_{c,m}^{i,j}(t), \nabla F_{c,m} \big( \bm{\theta}_{c,m}^{i,j}(t) \big) \rangle \big] \nonumber \\
    & \hspace{0.3cm} \overset{(b)}{\leq} \frac{2 \eta(t)}{MC} \sum_{c = 1}^{C} \sum_{m = 1}^{M} \sum_{i = 1}^{I} \sum_{j = 2}^{\tau} \mathbb{E}  \Big[ F_{c,m}(\bm{\theta}^{*}) - F_{c,m}(\bm{\theta}_{c,m}^{i,j}(t)) - \frac{\mu}{2} \big\| \bm{\theta}_{c,m}^{i,j}(t) - \bm{\theta}^{*} \big\|_{2}^{2} \Big] \nonumber \\
    & \hspace{0.3cm} = \frac{2 \eta(t)}{MC} \sum_{c = 1}^{C} \sum_{m = 1}^{M} \sum_{i = 1}^{I} \sum_{j = 2}^{\tau} \mathbb{E} \Big[ F_{c,m}(\bm{\theta}^{*}) - F_{c,m}^{*} + F_{c,m}^{*} - F_{c,m}(\bm{\theta}_{c,m}^{i,j}(t)) - \frac{\mu}{2} \big\| \bm{\theta}_{c,m}^{i,j}(t) - \bm{\theta}^{*} \big\|_{2}^{2} \Big] \nonumber \\
    & \hspace{0.3cm} = 2 \eta(t) I (\tau - 1) \Big( F^{*} - \frac{1}{MC} \sum_{c = 1}^{C} \sum_{m = 1}^{M} F_{c,m}^{*} \Big) + \frac{2 \eta}{MC}  \sum_{c = 1}^{C} \sum_{m = 1}^{M} \sum_{i = 1}^{I} \sum_{j = 2}^{\tau} \Big( F_{c,m}^{*} - \mathbb{E} \left[ F_{c,m}(\bm{\theta}_{c,m}^{i,j}) \right] \Big) \nonumber \\
    & \hspace{0.7cm} - \frac{\mu \eta(t)}{MC} \sum_{c = 1}^{C} \sum_{m = 1}^{M} \sum_{i = 1}^{I} \sum_{j = 2}^{\tau} \mathbb{E} \big[ \big\| \bm{\theta}_{c,m}^{i,j}(t) - \bm{\theta}^{*} \big\|_{2}^{2} \big] \nonumber \\
    & \hspace{0.3cm} \overset{(c)}{\leq} 2 \eta(t) I (\tau - 1) \Gamma - \frac{\mu \eta}{MC} \sum_{c = 1}^{C} \sum_{m = 1}^{M} \sum_{i = 1}^{I} \sum_{j = 2}^{\tau} \mathbb{E} \big[ \big\| \bm{\theta}_{c,m}^{i,j}(t) - \bm{\theta}^{*} \big\|_{2}^{2} \big], \label{eq: lemma2aproof1}
\end{align}
where (a) is obtained using $\mathbb{E}_{\xi} \left[ \nabla F_{c,m} \left( \bm{\theta}_{c,m}^{i,j}(t), \bm{\xi}_{c,m}^{i,j}(t) \right) \right] = F_{c,m}(\bm{\theta}_{c,m}^{i,j}(t)), \forall i,j,m,c,t$, (b) is using the $\mu$-strong convexity of $F_{c,m}$, (c) follows since $F_{c,m}^{*} \leq F_{c,m}(\bm{\theta}_{c,m}^{i,j}(t))$. Also
\begin{align}
    & \! - \! \left\| \bm{\theta}_{c,m}^{i,j}(t) \! - \! \bm{\theta}^{*} \right\|_{2}^{2} \! = \! - \! \left\| \bm{\theta}_{c,m}^{i,j}(t) \! - \! \bm{\theta}_{PS}(t) \right\|_{2}^{2} \! - \! \left\| \bm{\theta}_{PS}(t) \! - \! \bm{\theta}^{*} \right\|_{2}^{2} \! - \! 2 \langle \bm{\theta}_{c,m}^{i,j}(t) \! - \! \bm{\theta}_{PS}(t), \bm{\theta}_{PS}(t) \! - \! \bm{\theta}^{*} \rangle \nonumber \\
    & \hspace{0.3cm} \overset{(a)}{\leq} \! - \! \left\| \bm{\theta}_{c,m}^{i,j}(t) \! - \! \bm{\theta}_{PS}(t) \right\|_{2}^{2} \! - \! \left\| \bm{\theta}_{PS}(t) \! - \! \bm{\theta}^{*} \right\|_{2}^{2} \! + \! \frac{1}{\eta(t)} \left\| \bm{\theta}_{c,m}^{i,j}(t) \! - \! \bm{\theta}_{PS}(t) \right\|_{2}^{2} \! + \! \eta(t) \left\| \bm{\theta}_{PS}(t) \!  - \! \bm{\theta}^{*} \right\|_{2}^{2} \nonumber \\
    & \hspace{0.3cm} = - \left( 1 - \eta(t) \right) \left\| \bm{\theta}_{PS}(t) - \bm{\theta}^{*} \right\|_{2}^{2} + \left( \frac{1}{\eta(t)} - 1 \right) \left\| \bm{\theta}_{c,m}^{i,j}(t) - \bm{\theta}_{PS}(t) \right\|_{2}^{2}, \label{eq: lemma2aproof2}
\end{align} 
where (a) is due to Cauchy-Schwarz inequality. Plugging (\ref{eq: lemma2aproof1}) and (\ref{eq: lemma2aproof2}) concludes Lemma \ref{lemma2aproof}.
\ifCLASSOPTIONcaptionsoff
  \newpage
\fi


\bibliographystyle{IEEEtran}
\bibliography{references.bib}

\begin{thebibliography}{10}
\providecommand{\url}[1]{#1}
\csname url@samestyle\endcsname
\providecommand{\newblock}{\relax}
\providecommand{\bibinfo}[2]{#2}
\providecommand{\BIBentrySTDinterwordspacing}{\spaceskip=0pt\relax}
\providecommand{\BIBentryALTinterwordstretchfactor}{4}
\providecommand{\BIBentryALTinterwordspacing}{\spaceskip=\fontdimen2\font plus
\BIBentryALTinterwordstretchfactor\fontdimen3\font minus
  \fontdimen4\font\relax}
\providecommand{\BIBforeignlanguage}[2]{{%
\expandafter\ifx\csname l@#1\endcsname\relax
\typeout{** WARNING: IEEEtran.bst: No hyphenation pattern has been}%
\typeout{** loaded for the language `#1'. Using the pattern for}%
\typeout{** the default language instead.}%
\else
\language=\csname l@#1\endcsname
\fi
#2}}
\providecommand{\BIBdecl}{\relax}
\BIBdecl

\bibitem{aygun2022}
O.~{Ayg{\"u}n}, M.~Kazemi, D.~{G{\"u}nd{\"u}z}, and T.~M. Duman, ``Hierarchical
  {over-the-air} federated edge learning,'' in \emph{2022 IEEE International
  Conference on Communications (ICC)}, Seoul, South Korea, May 2022.

\bibitem{wei2020federated}
K.~Wei \emph{et~al.}, ``Federated learning with differential privacy:
  Algorithms and performance analysis,'' \emph{IEEE Transactions on Information
  Forensics and Security}, vol.~15, pp. 3454--3469, 2020.

\bibitem{lim2020federated}
W.~Y.~B. Lim \emph{et~al.}, ``Federated learning in mobile edge networks: A
  comprehensive survey,'' \emph{IEEE Commun. Surveys Tuts.}, vol.~22, no.~3,
  pp. 2031--2063, 2020.

\bibitem{mcmahan2017communication}
B.~McMahan, E.~Moore, D.~Ramage, S.~Hampson, and B.~A. y~Arcas,
  ``Communication-efficient learning of deep networks from decentralized
  data,'' in \emph{Artificial intelligence and statistics}.\hskip 1em plus
  0.5em minus 0.4em\relax PMLR, 2017, pp. 1273--1282.

\bibitem{gunduz2020communicate}
D.~G{\"u}nd{\"u}z \emph{et~al.}, ``Communicate to learn at the edge,''
  \emph{IEEE Commun. Mag.}, vol.~58, no.~12, pp. 14--19, 2020.

\bibitem{sery2021over}
T.~Sery, N.~Shlezinger, K.~Cohen, and Y.~C. Eldar, ``Over-the-air federated
  learning from heterogeneous data,'' \emph{IEEE Trans. Signal Process.},
  vol.~69, pp. 3796--3811, 2021.

\bibitem{liu2020privacy}
D.~Liu and O.~Simeone, ``Privacy for free: Wireless federated learning via
  uncoded transmission with adaptive power control,'' \emph{IEEE J. Sel. Areas
  Commun.}, vol.~39, no.~1, pp. 170--185, 2020.

\bibitem{chen2021joint}
M.~Chen, N.~Shlezinger, H.~V. Poor, Y.~C. Eldar, and S.~Cui, ``Joint resource
  management and model compression for wireless federated learning,'' in
  \emph{2021 IEEE International Conference on Communications (ICC)}, Montreal,
  Canada, Jun. 2021, pp. 1--6.

\bibitem{konevcny2016federated}
J.~Kone{\v{c}}n{\`y} \emph{et~al.}, ``Federated learning: Strategies for
  improving communication efficiency,'' \emph{arXiv preprint arXiv:1610.05492},
  2016.

\bibitem{zhang2021client}
W.~Zhang, X.~Wang, P.~Zhou, W.~Wu, and X.~Zhang, ``Client selection for
  federated learning with non-iid data in mobile edge computing,'' \emph{IEEE
  Access}, vol.~9, pp. 24\,462--24\,474, 2021.

\bibitem{zhao2018federated}
Y.~Zhao \emph{et~al.}, ``Federated learning with non-iid data,'' \emph{arXiv
  preprint arXiv:1806.00582}, 2018.

\bibitem{briggs2020federated}
C.~Briggs, Z.~Fan, and P.~Andras, ``Federated learning with hierarchical
  clustering of local updates to improve training on non-iid data,'' in
  \emph{2020 International Joint Conference on Neural Networks (IJCNN)},
  Glasgow, UK, Sep. 2020, pp. 1--9.

\bibitem{data2021byzantine}
D.~Data and S.~Diggavi, ``Byzantine-resilient sgd in high dimensions on
  heterogeneous data,'' in \emph{2021 IEEE Int'l Symp. on Inform. Theory
  (ISIT)}, Melbourne, Australia, Jul. 2021, pp. 2310--2315.

\bibitem{so2020byzantine}
J.~So, B.~G{\"u}ler, and A.~S. Avestimehr, ``Byzantine-resilient secure
  federated learning,'' \emph{IEEE J. Sel. Areas Commun.}, vol.~39, no.~7, pp.
  2168--2181, 2020.

\bibitem{dinh2020federated}
C.~T. Dinh \emph{et~al.}, ``Federated learning over wireless networks:
  Convergence analysis and resource allocation,'' \emph{IEEE/ACM Trans. on
  Networking}, vol.~29, no.~1, pp. 398--409, 2020.

\bibitem{luo2020hfel}
S.~Luo, X.~Chen, Q.~Wu, Z.~Zhou, and S.~Yu, ``Hfel: Joint edge association and
  resource allocation for cost-efficient hierarchical federated edge
  learning,'' \emph{IEEE Trans. Wireless Commun.}, vol.~19, no.~10, pp.
  6535--6548, 2020.

\bibitem{amiri2021convergence}
M.~M. Amiri, D.~G{\"u}nd{\"u}z, S.~R. Kulkarni, and H.~V. Poor, ``Convergence
  of update aware device scheduling for federated learning at the wireless
  edge,'' \emph{IEEE Trans. Wireless Commun.}, vol.~20, no.~6, pp. 3643--3658,
  2021.

\bibitem{sun2021dynamic}
Y.~Sun, S.~Zhou, Z.~Niu, and D.~G{\"u}nd{\"u}z, ``Dynamic scheduling for
  over-the-air federated edge learning with energy constraints,'' \emph{IEEE J.
  Sel. Areas Commun.}, vol.~40, no.~1, pp. 227--242, 2021.

\bibitem{amiri2021federated}
M.~M. Amiri, S.~R. Kulkarni, and H.~V. Poor, ``Federated learning with downlink
  device selection,'' in \emph{2021 IEEE 22nd International Workshop on Signal
  Processing Advances in Wireless Communications (SPAWC)}, Lucca, Italy, Sept.
  2021, pp. 306--310.

\bibitem{ren2020scheduling}
J.~Ren \emph{et~al.}, ``Scheduling for cellular federated edge learning with
  importance and channel awareness,'' \emph{IEEE Trans. Wireless Commun.},
  vol.~19, no.~11, pp. 7690--7703, 2020.

\bibitem{amiri2020machine}
M.~M. Amiri and D.~G{\"u}nd{\"u}z, ``Machine learning at the wireless edge:
  Distributed stochastic gradient descent over-the-air,'' \emph{IEEE Trans.
  Signal Process.}, vol.~68, pp. 2155--2169, 2020.

\bibitem{amiri2021blind}
M.~M. Amiri, T.~M. Duman, D.~G{\"u}nd{\"u}z, S.~R. Kulkarni, and H.~V. Poor,
  ``Blind federated edge learning,'' \emph{IEEE Trans. Wireless Commun.},
  vol.~20, no.~8, pp. 5129--5143, 2021.

\bibitem{zhu2019broadband}
G.~Zhu, Y.~Wang, and K.~Huang, ``Broadband analog aggregation for low-latency
  federated edge learning,'' \emph{IEEE Trans. Wireless Commun.}, vol.~19,
  no.~1, pp. 491--506, 2019.

\bibitem{shao2021federated}
Y.~Shao, D.~G{\"u}nd{\"u}z, and S.~C. Liew, ``Federated edge learning with
  misaligned over-the-air computation,'' \emph{IEEE Trans. Wireless Commun.},
  2021.

\bibitem{wei2022federated}
X.~Wei and C.~Shen, ``Federated learning over noisy channels: Convergence
  analysis and design examples,'' \emph{IEEE Trans. Cogn. Commun. Netw.}, 2022.

\bibitem{amiri2020federated}
M.~M. Amiri and D.~G{\"u}nd{\"u}z, ``Federated learning over wireless fading
  channels,'' \emph{IEEE Trans. Wireless Commun.}, vol.~19, no.~5, pp.
  3546--3557, 2020.

\bibitem{zhu2020one}
G.~Zhu, Y.~Du, D.~G{\"u}nd{\"u}z, and K.~Huang, ``One-bit over-the-air
  aggregation for communication-efficient federated edge learning: Design and
  convergence analysis,'' \emph{IEEE Trans. Wireless Commun.}, vol.~20, no.~3,
  pp. 2120--2135, 2020.

\bibitem{chen2021}
M.~Chen, N.~Shlezinger, H.~V. Poor, Y.~C. Eldar, and S.~Cui,
  ``Communication-efficient federated learning,'' \emph{Proceedings of the
  National Academy of Sciences}, vol. 118, no.~17, 2021.

\bibitem{abad2020hierarchical}
M.~S.~H. Abad, E.~Ozfatura, D.~Gunduz, and O.~Ercetin, ``Hierarchical federated
  learning across heterogeneous cellular networks,'' in \emph{ICASSP 2020-2020
  IEEE International Conference on Acoustics, Speech and Signal Processing
  (ICASSP)}, Barcelona, Spain, May 2020, pp. 8866--8870.

\bibitem{liu2020client}
L.~Liu, J.~Zhang, S.~Song, and K.~B. Letaief, ``Client-edge-cloud hierarchical
  federated learning,'' in \emph{ICC 2020-2020 IEEE International Conference on
  Communications (ICC)}, Dublin, Ireland, Jun. 2020, pp. 1--6.

\bibitem{liu2021hierarchical}
------, ``Hierarchical quantized federated learning: Convergence analysis and
  system design,'' \emph{arXiv preprint arXiv:2103.14272}, 2021.

\bibitem{wang2020local}
J.~Wang, S.~Wang, R.-R. Chen, and M.~Ji, ``Local averaging helps: Hierarchical
  federated learning and convergence analysis,'' \emph{arXiv preprint
  arXiv:2010.12998}, 2020.

\bibitem{mnist}
Y.~LeCun, ``The {MNIST} database of handwritten digits,'' \emph{http://yann.
  lecun. com/exdb/mnist/}, 1998.

\bibitem{cifar10}
A.~Krizhevsky \emph{et~al.}, ``Learning multiple layers of features from tiny
  images,'' 2009.

\bibitem{adam}
D.~P. Kingma and J.~Ba, ``Adam: A method for stochastic optimization,''
  \emph{arXiv preprint arXiv:1412.6980}, 2014.

\end{thebibliography}
%



%




\end{document}